\documentclass[letterpaper, 10 pt, conference]{ieeeconf}  % Comment this line out if you need a4paper

\IEEEoverridecommandlockouts                              % This command is only needed if 
\usepackage{tikz}
\usepackage{graphicx}
\usepackage{comment}
\usepackage{amsmath,amssymb} % define this before the line numbering.
\usepackage{color}
\usepackage{times}
\usepackage{epsfig}
\usepackage{subfigure}
\usepackage{algorithm}
\usepackage[noend]{algpseudocode}
\usepackage{multirow}
\usepackage{bm}
\usepackage{float}
\usepackage{booktabs}
\usepackage{bbding}
\usepackage{makecell}

\usepackage{times}
\usepackage{graphicx}
\usepackage[unicode]{hyperref}

\makeatletter
\newcommand\figcaption{\def\@captype{figure}\caption}
\newcommand\tabcaption{\def\@captype{table}\caption}
\makeatother

\usepackage[accsupp]{axessibility}  % Improves PDF readability for those with disabilities.

\DeclareMathOperator*{\argmin}{arg\,min}

\definecolor{MyGreen}{RGB}{30, 232, 114}
\definecolor{MyBlue}{RGB}{16, 83, 227}
\definecolor{MyRed}{RGB}{227, 16, 55}
\title{\LARGE \bf
AdaSfM: From Coarse Global to Fine Incremental Adaptive \\ Structure from Motion
}

\author{Yu Chen$^{1}$, Zihao Yu$^{2}$, Shu Song$^{2}$, Tianning Yu$^{3}$, Jianming Li$^{3}$, Gim Hee Lee$^{1}$ % <-this % stops a space
\thanks{${^1}$School of Computing, National University of Singapore, {\tt\small \{chenyu, gimhee.lee\}@comp.nus.edu.sg}}%
\thanks{${^2}$Segway-Ninebot Robotics Co., Ltd, {\tt\small yuzihao@buaa.edu.cn, songshu0905@gmail.com}}%
\thanks{${^3}$Navimow B.V. Co., Ltd, {\tt\small tianning.yu@rlm.segway.com, jianming.li@ninebot.com}}%
}
\begin{document}

\maketitle
\thispagestyle{empty}
\pagestyle{empty}

\vspace{-0.20in}
\begin{abstract}

Despite the impressive results achieved by many existing Structure from Motion (SfM) approaches, %achieved tremendous progress, 
there is still a need to improve the robustness, accuracy, and efficiency on large-scale scenes with many outlier matches and sparse view graphs. In this paper, we propose AdaSfM: a coarse-to-fine adaptive SfM approach that is scalable to large-scale and challenging datasets. Our approach first does a coarse global SfM which improves the reliability of the view graph by leveraging measurements from low-cost sensors such as Inertial Measurement Units (IMUs) and wheel encoders. Subsequently, the view graph is divided into sub-scenes that are refined in parallel by a fine local incremental SfM regularised by the result from the coarse global SfM to improve the camera registration accuracy 
and alleviate scene drifts. Finally, our approach uses a threshold-adaptive strategy to align all local reconstructions to the coordinate frame of global SfM. Extensive experiments on large-scale benchmark datasets show that our approach achieves state-of-the-art accuracy and efficiency.

\end{abstract}

%%%%%%%%%%%%%%%%%%%%%%%%%%%%%%%%%%%%%%%%%%%%%%%%%%%%%%%%%%%%%%%%%%%%%%%%%%%%%%%%%%%%%%%%%%%%%%%%%%%%%
\section{Introduction}

Structure from Motion (SfM) is an important topic that has been studied intensively over the past two decades. It has wide applications in augmented reality and autonomous driving for visual localization~\cite{DBLP:conf/cvpr/SarlinULGTLPLHK21,Brachmann_2021_ICCV,Dusmanu_2021_ICCV}, 
and in multi-view stereo~\cite{DBLP:journals/ftcgv/FurukawaH15,DBLP:conf/eccv/YaoLLFQ18} and novel view synthesis~\cite{DBLP:conf/eccv/MildenhallSTBRN20} 
by providing camera poses and optional sparse scene structures.

Despite the impressive results from many existing works, SfM remains challenging in two aspects. The first challenge is outlier feature matches caused by the diversity of scene features, e.g. texture-less, self-similar, non-Lambertian, etc. These diverse features impose challenges in sparse feature extraction and matching which result in outliers that are detrimental to the subsequent reconstruction process.
Incremental SfM~\cite{DBLP:journals/cacm/AgarwalFSSCSS11,DBLP:conf/cvpr/SchonbergerF16} is notoriously known to suffer from drift due to error accumulation, though is robust in handling outliers.
Global SfM methods~\cite{DBLP:conf/iccv/MoulonMM13,DBLP:conf/iccv/CuiT15,DBLP:conf/iccv/SweeneySHTP15} 
are proposed to handle drift, but fail to solve the scale ambiguities~\cite{DBLP:conf/cvpr/OzyesilS15} of camera 
positions and are not robust to outliers~\cite{DBLP:conf/cvpr/Govindu01,DBLP:conf/eccv/WilsonS14}. % Furthermore, existing translation averaging (TA) methods~\cite{DBLP:conf/cvpr/Govindu01,DBLP:conf/eccv/WilsonS14} are sensitive to outliers.

\begin{figure}[htbp]
  \centering
  \vspace{-0.12in}
  \includegraphics[width=0.98\linewidth]{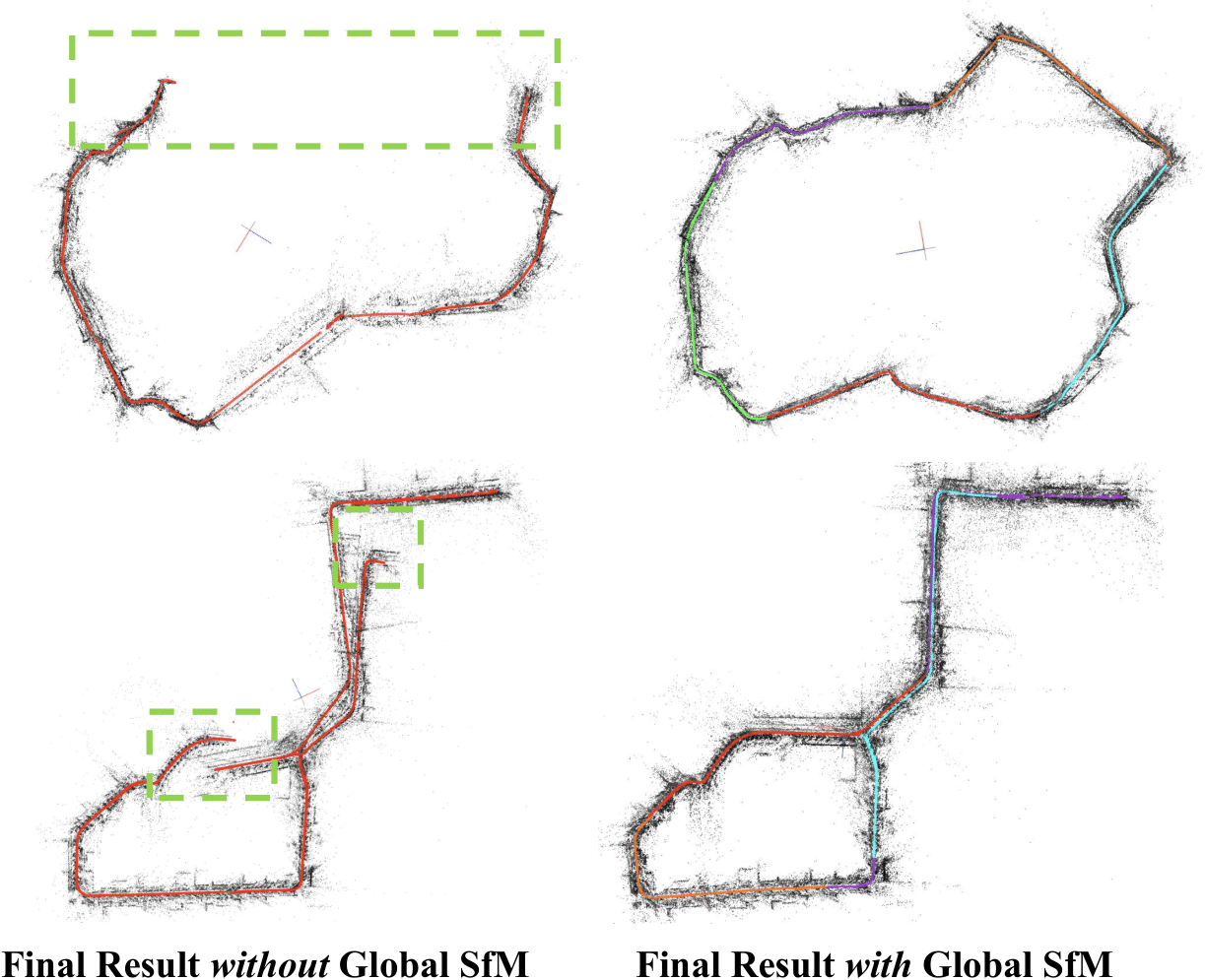}
  \vspace{-0.12in}

  \label{fig:4seasons_teaser}
  \caption{When combining with global SfM, our AdaSfM is more robust than traditional incremental SfM
           (tested on the public 4Seasons dataset~\cite{DBLP:conf/dagm/WenzelWYCKSZC20}).}
  \vspace{-0.26in}
\end{figure}

The second challenge is sparse view graphs from some large-scale datasets. Incremental SfM is known to be inefficient on large-scale datasets.
Several works \cite{DBLP:conf/accv/BhowmickPCGB14,DBLP:journals/pr/ChenSCW20,Zhu2017Parallel,DBLP:conf/cvpr/ZhuZZSFTQ18} have been proposed to handle %this challenge from thousands of images
% large-scale datasets with 
millions of images. These are divide-and-conquer SfM methods that deal with very large-scale datasets by grouping images into partitions. Each partition is processed by
a cluster of servers that concurrently circumvents the memory limitation. However, these methods
~\cite{DBLP:conf/accv/BhowmickPCGB14,DBLP:journals/pr/ChenSCW20,Zhu2017Parallel,DBLP:conf/cvpr/ZhuZZSFTQ18} 
are often limited to internet datasets or aerial images where the view graphs are very densely connected. The dense connections in the view graph ensure that there are sufficient constraints between the graph partitions.  
Nonetheless, divide-and-conquer methods often fail in datasets with weak 
associations between images for local reconstruction alignments or lack of visual constraints for stable camera registration. An example of such a dataset is autonomous self-driving cars where the interval between consecutive images can be large. 

% In wild scenes, such as the scenarios where autonomous driving cars run, divide-and-conquer methods often fail, due to weak associations between images 
% for local reconstructions alignment or lack of visual constraints for stable camera registration.

% \begin{figure}[htbp]
%   \centering
%     \includegraphics[width=0.95\linewidth]{img/teaser_jy1_global_to_ada.pdf}

%   \caption{\textbf{Reconstruction results from coarse global SfM to finer parallel incremental SfM}.
%   The left image is obtained by our initial global SfM method, the right image is the refined
%   result by leveraging the poses from previous global SfM.}
%   \label{fig:teaser_jy1_global_to_ada}
% \end{figure}

\begin{figure*}[htbp]
  \centering
    \includegraphics[width=0.95\linewidth]{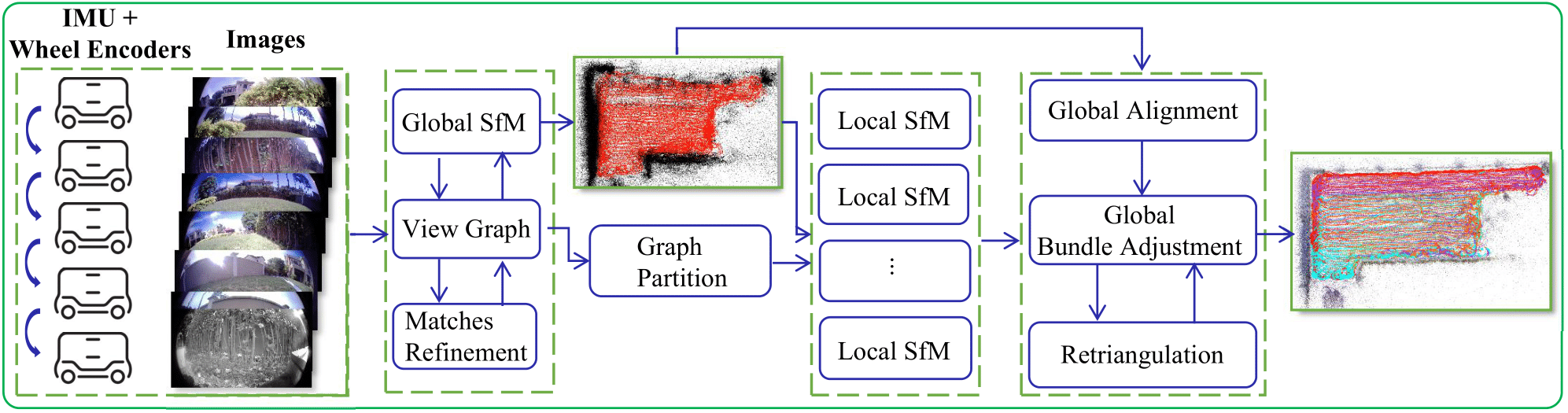}
  
  \vspace{-0.16in}
  \caption{\textbf{The pipeline of our proposed SfM method}. Our method takes images and measurements from low-cost sensors as inputs. The view graph is built after feature matching and refined by the 
           result of global SfM. The absolute poses from the global SfM are used as priors in the subsequent local SfM process. The final reconstruction result is merged into the global SfM reference frame.}
  \label{fig:pipeline}
  \vspace{-0.2in}
\end{figure*}

In view of the challenges from the outlier feature matches and sparse view graphs %large-scale datasets
on the existing SfM approaches, we propose AdaSfM: a coarse-to-fine adaptive SfM pipeline to enhance the robustness of SfM in dealing 
with large-scale challenging scenes. Specifically, we first solve the global SfM at a coarse scale, and then the 
result of the global SfM is used to enhance the scalability of the local incremental reconstruction. Both the scale ambiguities and outlier ratio in global SfM can be significantly reduced by incorporating measurements from the IMU and wheel encoder, which are often available in mobile devices or 
autonomous self-driving cars.
We preintegrate~\cite{DBLP:conf/rss/ForsterCDS15} the IMU measurements to get the relative 
poses of consecutive frames $\mathcal{P}_{t}=\{\mathbf{P}_{t_0}, \mathbf{P}_{t_1}, \cdots\}$, and use the measurements from the wheel 
encoder to constrain scale drifts of the IMU preintegration~\cite{DBLP:conf/icra/WuGGR17}. We then replace the relative poses of the consecutive frames in the view graph formed by two-view geometry~\cite{DBLP:books/cu/HZ2004,DBLP:conf/cvpr/SchonbergerF16} with $\mathcal{P}_{t}$ estimated by the IMU and wheel encoder. 
This augmented view graph is then used to estimate the global poses. Consequently, we obtain a coarse scene structure and camera poses, where the latter can be used to filter wrong feature matches.
%  The global poses are used to filter wrong 
% feature matches in the view graph, and we further utilize the global poses to estimate more accurate poses in a divide-and-conquer manner. 
Since that, we partition the view graph with the existing graph cut method~\cite{DBLP:journals/pami/ShiM00} and then extend the sub-graphs with a novel adaptive flood-fill method to 
enhance the constraints of separators~\cite{journals/siam/Lipton1979}. We define separators as images that connect different sub-graphs. For each local SfM, the poses from the global SfM are used for camera registration and to constrain the global refinement of 3D points and camera poses. Finally, we design an adaptive global alignment 
strategy to merge local reconstructions with the coordinate frame of the global SfM set as the reference frame.
% An overview of the effectiveness of our method can be seen in Fig.~\ref{fig:4seasons_teaser}.
We illustrate the pipeline of our method in Fig.~\ref{fig:pipeline}.
% Fig.~\ref{fig:teaser_jy1_global_to_ada} shows the results obtained by our method, which 
% is reconstructed from a dataset with more than 30,000 images.

% Our main contributions to this work are:
% \begin{itemize}
%   \item We fused relative motions from low-cost sensors to improve the solvability of the view graph and made global 
%         SfM is practical in challenging scenes.
%   \item We proposed a flood-fill graph partition algorithm to enhance the constraints for separators.
%   \item We utilized the poses from global SfM to improve the robustness of local incremental SfM.
%   \item We proposed an adaptive alignment algorithm to robustly merge local reconstruction results.
% \end{itemize}

We evaluate our method extensively on large-scale challenging 
scenes. Experimental results show that our AdaSfM is adaptive to different scene structures. Furthermore, we achieve better robustness and  
comparable efficiency in comparison to existing state-of-the-art SfM methods.

%---------------------------------------------------------------------------------------------
\section{Related Work}
\label{sec:related_work}

\noindent \textbf{Incremental SfM.}
Agarwal \emph{et al.} \cite{DBLP:journals/cacm/AgarwalFSSCSS11} apply preconditioned conjugate gradient~\cite{DBLP:conf/eccv/AgarwalSSS10} 
to accelerate large-scale BA~\cite{DBLP:conf/iccvw/TriggsMHF99}.
The drift problem is alleviated in~\cite{DBLP:conf/3dim/Wu13}
with a re-triangulation (RT) step before global BA. Sch{\"{o}}nberger and Frahm 
\cite{DBLP:conf/cvpr/SchonbergerF16} augment the view graph by estimating multiple geometric 
models in geometric verification and improve the image registration robustness with next best view selection. 
In addition to the RT before BA~\cite{DBLP:conf/3dim/Wu13}, RT is also performed after BA in~\cite{DBLP:conf/cvpr/SchonbergerF16}.
To reduce the time complexity of repetitive image registration, Cui \emph{et al}~\cite{DBLP:conf/3dim/CuiSGH17} select 
a batch of images for registration, and select a subset of good tracks for BA.

\noindent \textbf{Global SfM.}
The simplest configuration of a global SfM method only requires 1) estimating the global rotations by rotation averaging (RA), 2) obtaining the global positions 
by TA, and 3) triangulating 3D points and performing a final global BA.
% Most global SfM methods~\cite{DBLP:conf/cvpr/Govindu01,DBLP:conf/cvpr/Govindu04} focus on solving the RA and TA problems.
Govindu~\cite{DBLP:conf/cvpr/Govindu04} represents rotations by lie-algebra, and global rotations and global positions are estimated simultaneously. Chatterjee and 
Govindu~\cite{DBLP:conf/iccv/ChatterjeeG13,DBLP:journals/pami/ChatterjeeG18} improve the rotation estimation of ~\cite{DBLP:conf/cvpr/Govindu04} by a robust $l_1$ initialization followed by a refinement of the rotations with iteratively reweighted least-squares (IRLS)~\cite{CSTM_Holland1977}.
To solve the TA problem, Wilson \emph{et al}~\cite{DBLP:conf/eccv/WilsonBS16} project relative translations onto the 1D space to identify outliers. Relative translations that are inconsistent with the translation directions that have the highest consensus are removed. A nonlinear least-squares problem is then solved to get the global positions. 
Goldstein \emph{et al.}~\cite{DBLP:conf/eccv/GoldsteinHLVS16} relax the scale constraints of
~\cite{DBLP:conf/eccv/WilsonBS16} to linear scale factors, and the convex linear programming problem is solved by 
ADMM~\cite{DBLP:journals/ftml/BoydPCPE11}. {\"{O}}zyesil and Singer~\cite{DBLP:conf/cvpr/OzyesilS15} utilize 
the parallel rigidity theory to select the images where positions can be estimated uniquely and solved as a constrained 
quadratic programming problem. % by IRLS.
By minimizing the $\sin \theta$ between two relative translations, 
Zhuang \emph{et al.}~\cite{DBLP:journals/cvpr/ZhuangCL18} improve the insensitivity to narrow baselines of 
TA. The robustness of TA is also improved in~\cite{DBLP:journals/cvpr/ZhuangCL18} by incorporating global rotations.
% from RA in the formulation.

\noindent \textbf{Hybrid SfM.}
Cui \emph{et al.} \cite{DBLP:conf/cvpr/CuiGSH17} obtain orientations by RA and 
then register camera centers incrementally with the perspective-2-point (P2P) algorithm.
Bhomick \emph{et al.}~\cite{DBLP:conf/accv/BhowmickPCGB14} propose to divide the scene graph, where 
the graph is built from the similarity scores between images. Feature matching and local SfM can then be 
executed in parallel and local reconstructions are merged~\cite{DBLP:conf/accv/BhowmickPCGB14}.
Zhu \emph{et al.} \cite{Zhu2017Parallel,DBLP:conf/cvpr/ZhuZZSFTQ18} adopt a similar strategy to divide the scene and 
the graph is constructed after feature matching. The relative poses are collected after merging all 
local incremental reconstruction results. The outliers are filtered during local reconstruction, global 
rotations are fixed by RA, and camera centers are registered with TA at 
the cluster level. Based on \cite{Zhu2017Parallel}, Chen \emph{et al.} \cite{DBLP:journals/pr/ChenSCW20} find the minimum 
spanning tree (MST) to solve the final merging step. The MST is constructed at the cluster level, and the most accurate 
similarity transformations between clusters are given by the MST. Locher \emph{et al.} \cite{DBLP:conf/eccv/LocherHG18} 
filtered wrong epipolar geometries by RA before applying the divide-and-conquer method~\cite{Zhu2017Parallel}.
Jiang \emph{et al.}~\cite{DBLP:conf/icit2/JiangTMO21} use a visual-inertial navigation system (VINS)
~\cite{DBLP:journals/trob/QinLS18} to first estimate the camera trajectories with loop detection and loop closure
~\cite{DBLP:journals/trob/Mur-ArtalMT15}. Images are then divided into sequences according to timestamps. 
However, \cite{DBLP:conf/icit2/JiangTMO21} requires two carefully designed systems: one 
for VINS with loop detection and the other for SfM. Loop detection is also a challenge in real-world scenes.

\section{Notations}
\label{sec:notation}

We denote the absolute camera poses as $\mathcal{P}=\{\mathbf{P}_i = [\mathbf{R}_i | \mathbf{t}_i] \}$, where
$\mathbf{R}_i, \mathbf{t}_i$ are the rotation and translation of the $i$-th image, respectively.
The absolute camera poses project 3D points $\mathcal{X}=\{\mathbf{X}_k\}$ from the world frame to the camera frame. The camera centers are denoted by $\{\mathbf{C}_i\}$. The relative pose from image $i$ to image $j$ are denoted as $\mathbf{P}_{ij}=[\mathbf{R}_{ij} | \mathbf{t}_{ij}]$, where 
$\mathbf{R}_{ij}, \mathbf{t}_{ij}$ are the relative rotations and translations, respectively.
We define the view graph as $\mathcal{G} = \{\mathcal{V}, \mathcal{E} \}$, where
$\mathcal{V}$ denotes the collection of images and $\mathcal{E}$ denotes the two view 
geometries, i.e. the relative poses and inlier matches between the image pairs. For two rotations 
$\mathbf{R}_i, \mathbf{R}_j$, we use $\log(\mathbf{R}_i, \mathbf{R}_j) = \log(\mathbf{R}_j \mathbf{R}_i^\top)$ 
to denote the angular error and $\| \mathbf{R}_i - \mathbf{R}_j \|_F$ to denote the chordal distance. Additionally, the keypoints and the normalized keypoints after applying the intrinsic 
matrix $\mathbf{K}$ are denoted by $\mathbf{u}$ and $\mathbf{\hat{u}}$, respectively.

%---------------------------------------------------------------------------------------------
\section{Coarse Global to Fine Incremental SfM}
\label{sec:global_sfm_to_finer_incremental}

In this section, we introduce our method in detail. In Sec.~\ref{subsec:coarse_global_sfm}, we introduce
our global SfM that can effectively cope with outliers in challenging scenes. A refinement step 
is also introduced to remove outlier matches after global SfM. In Sec.~\ref{subsec:finer_paralle_incremental_sfm},
we describe our parallel incremental SfM approach that utilizes the results from coarse global 
SfM to mitigate the problems from sparse view graphs.

% \vspace{-0.4in}
\subsection{Coarse Global SfM}
\label{subsec:coarse_global_sfm}

We first obtain the absolute rotations $\mathbf{R}_i$ by solving the rotation averaging problem:
\begin{small}
\begin{equation}
\label{equ:rotation_averaging}
  \argmin_{\{ \hat{\mathbf{R}}_i \}}\ \sum_{i \in \mathcal{V}, \atop (i,j) \in \mathcal{E}} 
    d (\hat{\mathbf{R}}_j \hat{\mathbf{R}}_i^\top, \mathbf{R}_{ij}),
\end{equation}
\end{small}
where $\hat{\mathbf{R}}_i$ denotes the absolute poses obtained by rotation averaging, and 
$d(\cdot)=\|\cdot\|_F$ denotes the chordal distance. Eq.~\eqref{equ:rotation_averaging} can be 
solved robustly and efficiently by~\cite{DBLP:conf/cvpr/Chen0K21}.
We then obtain the absolute camera positions by solving the translation averaging problem.
However, existing translation averaging methods often fail to recover the camera positions 
under challenging scenes due to two main factors: 1) The high ratio of outliers in the relative translations. 2) The view graph is solvable only when the parallel rigid graph condition~\cite{DBLP:conf/cvpr/OzyesilS15} is satisfied.
To alleviate the first problem, we first remove the erroneous matching pairs by
checking the discrepancy of relative rotations:
$\log (\mathbf{R}_{ij}^\top \hat{\mathbf{R}}_j \hat{\mathbf{R}}_i^\top) > \epsilon_{\mathbf{R}}$, and then the relative translations~\cite{DBLP:conf/cvpr/OzyesilS15} are refined in parallel by:
\begin{small}
\begin{equation}
  \argmin_{\mathbf{t}_{ij}}\ 
    \| \hat{\mathbf{u}}'^{\top} ([\mathbf{t}_{ij}]_{\times} (\hat{\mathbf{R}}_j \hat{\mathbf{R}}_i^\top)) \hat{\mathbf{u}} \|, \quad\text{s.t.}\quad \|\mathbf{t}_{ij}\|=1.
\end{equation}
\end{small}

We do not extract the rigid parallel graph~\cite{DBLP:conf/cvpr/OzyesilS15} 
to solve the scale ambiguities since it is time-consuming to solve polynomial equations. Furthermore, the state-of-the-art method to 
establish the solvability of a view graph is only limited to 90 nodes~\cite{Arrigoni_2021_ICCV}. We improve the solvability of the view graph by augmenting the relative translations in $\mathcal{P}_t$ of the consecutive frames from the IMU and wheel encoder. We do not augment the relative rotations because they are more accurate from the image-based two-view geometry.
% instead augment the view graph by the fused relative poses. 
% to 
% Specifically, for the relative translations that already existed in the view graph, we replace them with the fused 
% relative translations. Moreover, we also add more fused relative poses to the view graph to alleviate the scale 
% ambiguities. 
Note that errors can accumulate increasingly in the augmented relative poses during the motion of the devices due to the bias of the accelerometers and gyroscopes in the IMU, and drifts in the wheel encoder caused by friction and wheel slippages. To circumvent this problem, we only use the relative poses where the time difference is below a threshold $\epsilon_{T}$.

\begin{figure}[htbp]
  \centering
  \begin{minipage}[t]{0.48\textwidth}
    \centering
    \includegraphics[width=0.98\linewidth]{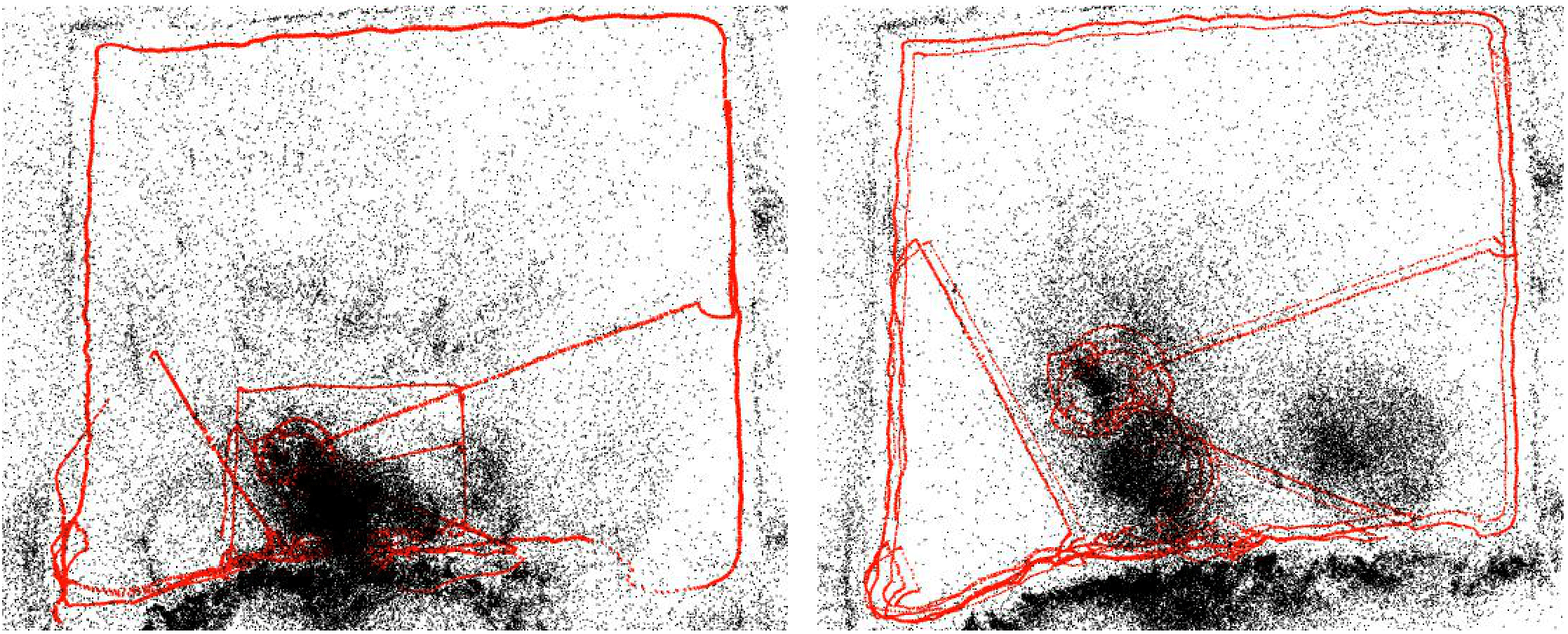}
    \caption{Comparison of global SfM results. Results from ~\cite{DBLP:conf/cvpr/OzyesilS15} (left) and Eq.~\eqref{equ:aug_translation_averaging} (right).
        Red and black colors respectively denote vehicle trajectories and sparse point clouds.}
    \label{fig:yht_compare}
  \end{minipage}
%   \quad
%   \begin{minipage}[t]{0.48\textwidth}
%     \centering
%     \includegraphics[width=1.0\linewidth]{img/711_compare}
%     \caption{\textbf{Effectiveness of matches refinement}. Results from
%              COLMAP~\cite{DBLP:conf/cvpr/SchonbergerF16} (left) and our method (right) after matches refinement.}
%     \label{fig:711_compare}
%   \end{minipage}
%   \vspace{-1cm}
\end{figure}

Since we obtained the \textit{augmented view graph} 
$\mathcal{G}_{\text{aug}}=\{\mathcal{V}, \mathcal{E}_{\text{aug}} \}$, the rigidity of the original view graph 
is augmented and the scale ambiguities of some images can be eliminated. We can then further solve the translation averaging problem below:
% \begin{small}
% \begin{align}
% \label{equ:aug_translation_averaging}
%    \arg\min\ &\sum_{(i, j) \in \mathcal{E}} \| s_{ij}(\hat{\mathbf{C}}_j - \hat{\mathbf{C}}_i) - \mathbf{R}_j \mathbf{t}_{ij} \| +
%               \sum_{(i, j) \in \mathcal{E}_{\text{aug}}, \atop (i, j) \notin \mathcal{E}} \| \hat{\mathbf{C}}_j - \hat{\mathbf{C}}_i - \mathbf{R}_j \mathbf{t}_{ij} \|, \\ \nonumber
%    \text{s.t.}\qquad & s_{ij} \geq 0\quad \forall (i, j) \in \mathcal{E},\quad \sum_{i \in \mathcal{V}} \hat{\mathbf{C}}_i = 0.
% \end{align}
% \end{small}
\begin{small}
\begin{align}
\label{equ:aug_translation_averaging}
     \argmin_{\substack{\hat{\mathbf{C}}_i, i \in \mathcal{V}; \atop ~s_{ij}, (i,j) \in \mathcal{E}_{\text{aug}}}} &\sum_{(i,j) \in \mathcal{E}_{\text{aug}}} \| s_{ij}(\hat{\mathbf{C}}_i - \hat{\mathbf{C}}_j) - \mathbf{R}_j^{\top} \mathbf{t}_{ij} \|, \\ \nonumber
     \text{s.t.}\qquad & s_{ij} \geq 0, \quad \forall (i,j) \in \mathcal{E}_{\text{aug}}; \quad \sum_{i \in \mathcal{V}} \hat{\mathbf{C}}_i = 0.
\end{align}
\end{small}
\eqref{equ:aug_translation_averaging} can be solved efficiently and
robustly under the $l_1$-norm by collecting all the constraints. 
% We demonstrate that Eq.~\eqref{equ:aug_translation_averaging} has 
% the following advantages: (1) The outlier ratio is reduced, as 
% some bad relative translations are corrected by the fused relative translations; (2) The scale ambiguities 
% of some images are eliminated, thus we can recover more camera positions uniquely. 
% Although we can use the absolute scales as 
% constraints during optimization for the edges 
% $\{(i, j) \mid (i, j) \in \mathcal{E}_{\text{aug}}, (i, j) \notin \mathcal{E} \}$, we find that there is no improvement in practice over the one without absolute scale constraints.
Note all the relative translations are normalized in $\mathcal{E}_{\text{aug}}$.
The right of Fig.~\ref{fig:yht_compare} shows our global SfM result by solving~\eqref{equ:aug_translation_averaging}.

After translation averaging, we triangulate the 3D points and perform an iterative global bundle adjustment 
to refine camera poses. It is worth mentioning that, global SfM can generate more tracks than incremental SfM, as 
its camera poses are less accurate and thus it fails to merge some tracks that are physically the same. Besides, according to~\cite{DBLP:conf/3dim/CuiSGH17}, 
tracks are redundant for optimisation. Therefore, we can reduce the computation and memory burden with fewer tracks.
Though a well-designed algorithm may help with the selection of tracks, we simply create tracks with a stricter threshold: only when the angle between the 
two rays respectively go through the 3D point and the two camera centers are larger than 5 degrees, it is deemed as a valid track.
Note that for numerical stability during optimization, the coordinates are normalized after each iteration.

\subsubsection{Matches Refinement}
\label{subsubsec:matches_refinement}
% By incorporating relative poses from low-cost sensors, global SfM can recover camera poses 
% correctly while discarding wrong two-view geometries. 
The correct camera poses recovered by our global SfM with the relative poses from the low-cost sensors to eliminate the wrong two-view geometry estimates can be further utilized to filter out wrong image feature matches. 
For a calibrated camera with known intrinsics, we can recover the essential matrix between images i and j from
$\hat{\mathbf{E}} = [\hat{\mathbf{t}}_{ij}]_{\times} \hat{\mathbf{R}}_{ij}$ 
with the absolute rotations $\hat{\mathbf{R}}_i$ and translations 
$\hat{\mathbf{t}}_i$ computed from rotation and translation averaging.
$(\hat{\mathbf{t}}_{ij}, \hat{\mathbf{R}}_{ij})$ are computed from $(\hat{\mathbf{R}}_i, \hat{\mathbf{R}}_j)$ 
and $(\hat{\mathbf{t}}_i, \hat{\mathbf{t}}_j)$. The true matches $\hat{\mathbf{u}}' \leftrightarrow \hat{\mathbf{u}}$ must satisfy the check on the total point-to-epipolar line distance~\cite{DBLP:books/cu/HZ2004} over the two views, i.e.
\vspace{-0.08in}
\begin{equation}
  \label{equ:epipolar_geometry_constraint}
%   \hat{\mathbf{u}}'^{\top} \hat{\mathbf{E}} \hat{\mathbf{u}} = 0.
    d_{\bot}(\hat{\mathbf{u}}, \mathbf{E}\hat{\mathbf{u}}') + d_{\bot}(\hat{\mathbf{u}}', \mathbf{E}\hat{\mathbf{u}}) \leq \epsilon_{M}.
\end{equation}
% \vspace{-0.1in}
$d_{\bot}(\mathbf{x},\mathbf{l})$ gives the shortest distance between a point $\mathbf{x}$ and a line $\mathbf{l}$. 
The epipolar lines on the two images are given by $\mathbf{l} = \mathbf{E}\hat{\mathbf{u}}'$ and $\mathbf{l}' = \mathbf{E}\hat{\mathbf{u}}$. $\epsilon_{M}$ is the threshold for the check. 

\begin{figure}[htbp]
        \centering
         \begin{minipage}[t]{0.52\textwidth}
          \vspace{0pt} 
          \begin{minipage}[t]{0.5\textwidth}
            \centering
            \includegraphics[width=1\linewidth]{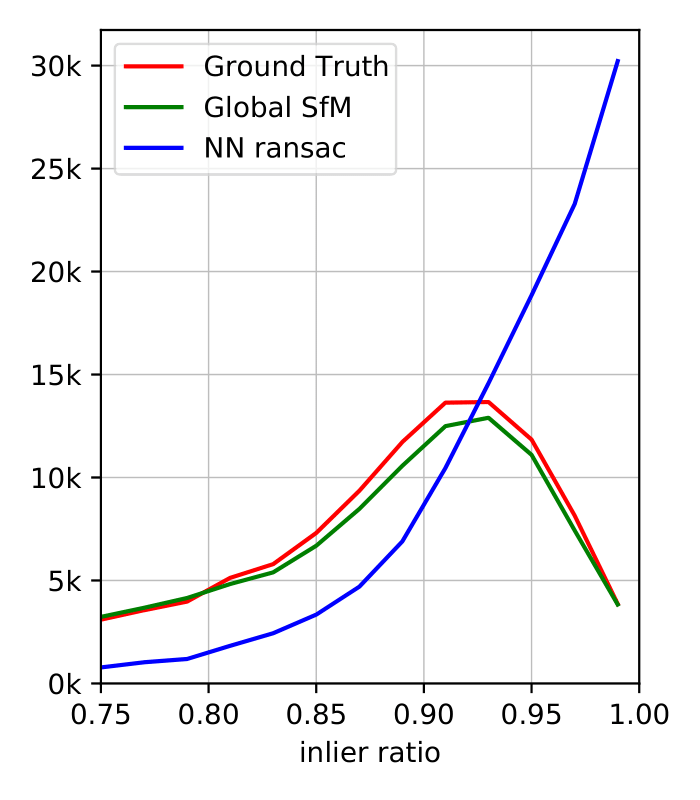}
          \end{minipage}%\quad
          \hfill
          \begin{minipage}[t]{0.5\textwidth}
          \centering
          \includegraphics[width=1\linewidth]{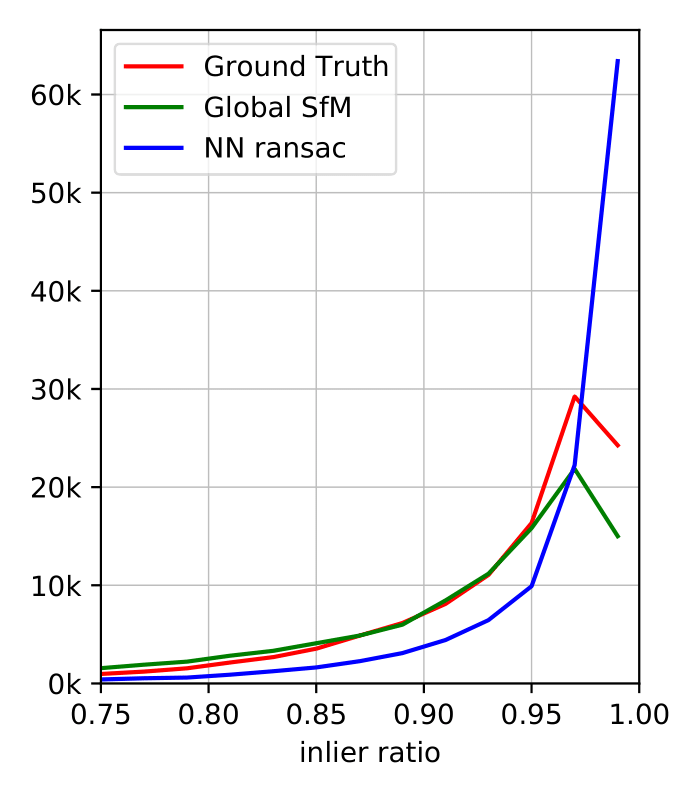}
        \end{minipage}%\quad
        \vspace{-0.22in}
        \figcaption{\textbf{Inlier ratio distribution} of NN+RANSAC, global SfM and \\
          incremental SfM (ground truth) on the 711 (left) and B6 (right) datasets.}
        \label{fig:inlier_ratio_hist}
        %  \hfill
      \end{minipage}%\quad
\vspace{-0.3in}
\end{figure}

The effectiveness of global SfM to filter wrong matches can be seen in Fig.~\ref{fig:match_refine_B6}. %~\ref{fig:711_compare}.
We build a pseudo ground truth by COLMAP~\cite{DBLP:conf/cvpr/SchonbergerF16} to evaluate the accuracy of the global SfM.
The ratio test is performed after NN by default. 
Fig.~\ref{fig:inlier_ratio_hist} shows the inlier ratio distribution after NN+RANSAC and matches refinement with 
relative poses obtained from global SfM and incremental SfM, respectively. 
Table.~\ref{table:auc_rel_poses} gives the relative pose estimation AUC of NN+RANSAC and global SfM with respect to incremental SfM. 
It can be seen that our coarse global SfM can obtain comparable accuracy to 
COLMAP~\cite{DBLP:conf/cvpr/SchonbergerF16} in the refinement of the matches.

\vspace{-0.1in}

\begin{table}[htbp]
\centering
\resizebox{0.5\textwidth}{!}{
  \begin{tabular}{c c c c c c c c c} %表格列
    \toprule

    \multirow{2}{*}{AUC}  &
    \multicolumn{2}{c}{\textbf{NN+RANSAC}}  & % B6
    \multicolumn{2}{c}{\textbf{Global SfM}} \quad& % B6
    \multicolumn{2}{c}{\textbf{NN+RANSAC}}  & % 711
    \multicolumn{2}{c}{\textbf{Global SfM}}\ \\ % 711
    \cmidrule(r){2-5} \cmidrule(r){6-9}
    % \cmidrule(r){2-3} \cmidrule(r){4-5} \cmidrule(r){6-7} \cmidrule(r){8-9}
      & \makecell[c]{$\mathbf{R}$} & \makecell[c]{$\mathbf{t}$}
      & \makecell[c]{$\mathbf{R}$} & \makecell[c]{$\mathbf{t}$} 
      & \makecell[c]{$\mathbf{R}$} & \makecell[c]{$\mathbf{t}$} 
      & \makecell[c]{$\mathbf{R}$} & \makecell[c]{$\mathbf{t}$} \\
    \midrule

    $\text{@} 0.1^{\circ}$  & 1.52 & 0.01 & 6.67 & 0.02 & 2.14 & 0.01 & 8.41 & 0.09    \\
    % $0.3^{\circ}$  & ---- & ---- & ---- & ---- & ---- & ---- & ---- & ----   \\
    $\text{@} 0.5^{\circ}$  & 14.74 & 0.25 & 44.87 & 0.48 & 21.47 & 0.36 & 44.14 & 1.96   \\
    $\text{@} 1.0^{\circ}$  & 28.92 & 0.96 & 64.15 & 1.80 & 40.99 & 1.40 & 64.48 & 6.48   \\
    $\text{@} 3.0^{\circ}$  & 55.75 & 5.85 & 84.76 & 9.60 & 68.08 & 9.18 & 86.89 & 24.00   \\
    $\text{@} 5.0^{\circ}$  & 68.27 & 10.94 & 90.34 & 17.71 & 77.39 & 17.58 & 92.06 & 35.41 \\
    $\text{@} 10.0^{\circ}$ & 81.71 & 20.21 & 94.99 & 33.03 & 86.81 & 32.46 & 96.01 & 51.07 \\
    $\text{@} 20.0^{\circ}$ & 90.29 & 29.97 & 97.48 & 49.87 & 92.90 & 46.95 & 98.00 & 64.65 \\

    \bottomrule
  \end{tabular}
}
\tabcaption{\textbf{Relative pose estimation AUC} of NN+RANSAC and global SfM with respect to
         incremental SfM on the B6 (column 2-5) and 711 (column 6-9) datasets.}
\label{table:auc_rel_poses}
\vspace{-0.4in}
\end{table}

% \vspace{-0.2in}
\subsection{Finer Parallel Incremental SfM}
\label{subsec:finer_paralle_incremental_sfm}

Although we have obtained the absolute camera poses by global SfM, %however,
these coarse poses are 
not accurate enough for localization. To improve the accuracy, we propose to refine the camera poses and scene structure with the divide-and-conquer incremental SfM.

\subsubsection{Adaptive Graph Partition}
\label{subsubsec:graph_partition}

\begin{figure*}[htbp]
  \centering
  % \subfigure[The initial view graph.]
  % {
  %   \begin{minipage}{0.4\linewidth}
  %     \label{fig:flood_fill_1}
  %     \centering
  %     \includegraphics[width=0.85\linewidth]{img/flood_fill_1}
  %   \end{minipage}
  % }\quad
  \subfigure[Initial graph cut.]
    %Separators are marked green, edges intersect with red lines are removed by the cut.]
  {
    \begin{minipage}{0.31\linewidth}
      \label{fig:flood_fill_2}
      \centering
      \includegraphics[width=0.85\linewidth]{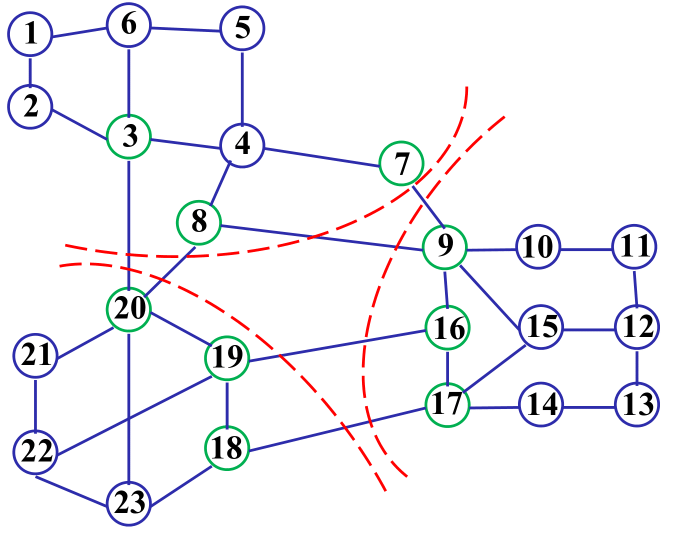}
    \end{minipage}
  }
  \subfigure[The 1st graph expansion.]
  {
    \begin{minipage}{0.29\linewidth}
      \label{fig:flood_fill_3}
      \includegraphics[width=0.85\linewidth]{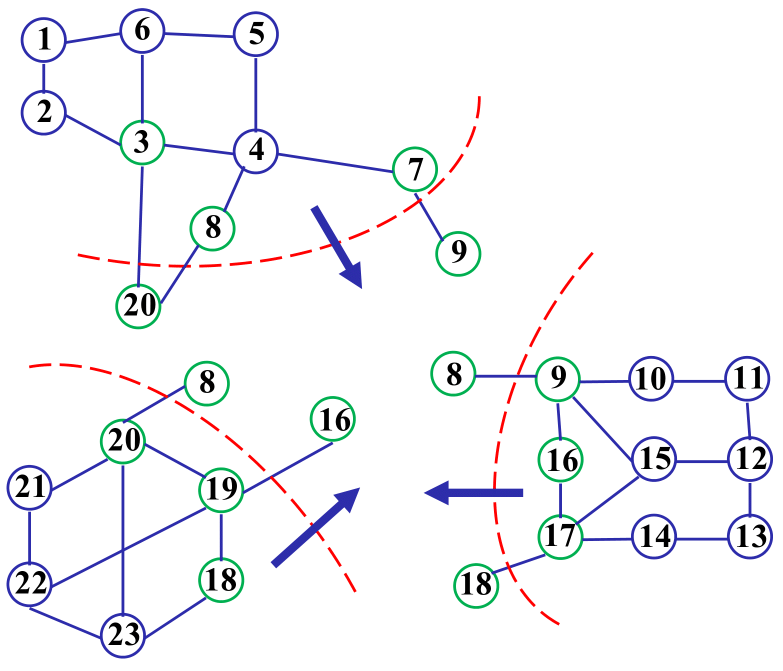}
    \end{minipage}
  }
  \subfigure[The 2nd graph expansion.]
  {
    \begin{minipage}{0.30\linewidth}
      \label{fig:flood_fill_4}
      \includegraphics[width=0.9\linewidth]{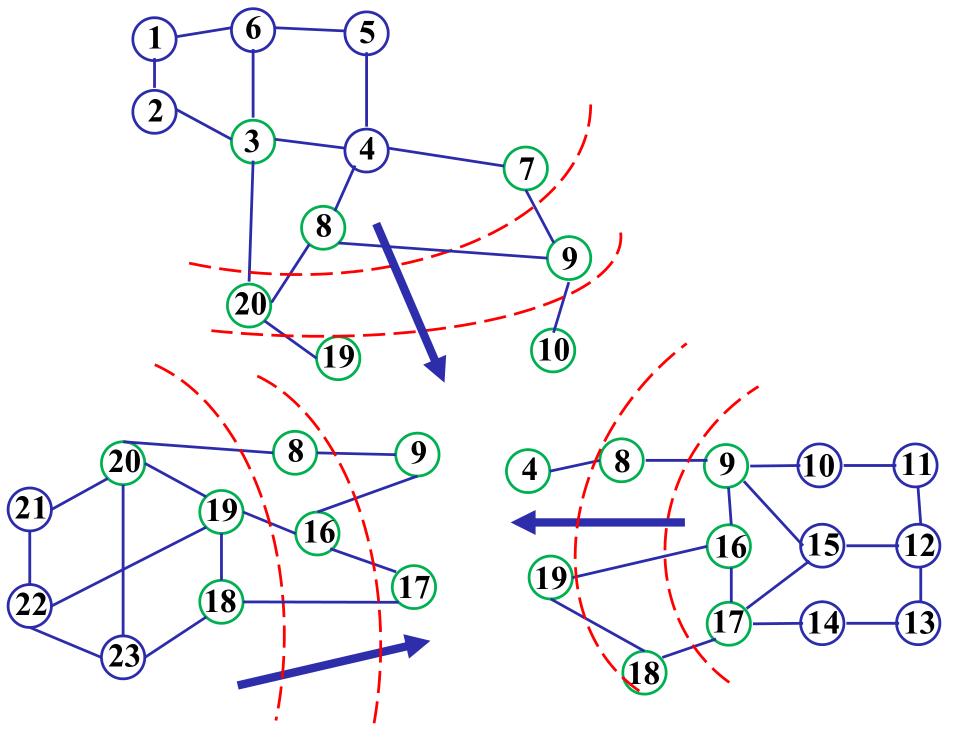}
    \end{minipage}
  }

  \vspace{-0.1in}
  \caption{\textbf{Pipeline of adaptive flood-fill graph partition}. In the view graph, nodes are denoted 
           by blue circles, edges are denoted by blue solid lines. Separators are marked by green circles.}
  \label{fig:flood_fill_graph_partition}
  \vspace{-0.18in}
\end{figure*}

Existing approaches~\cite{Zhu2017Parallel,DBLP:journals/pr/ChenSCW20} used a cut-and-expand schema to create overlapping areas between partitions. 
However, these approaches have two main drawbacks: 
: 1) The overlapping areas are not enough for final merging when the view graph becomes too sparse. This can
be seen from Fig.~\ref{fig:flood_fill_2}. Edges (3, 20), (7, 9), (8, 9), (8, 20), (16, 19), (17, 18)
are collected after the graph cut, and then the images on these edges are added as separators of the partitions. In Fig.~\ref{fig:flood_fill_2}, only images
$\{3, 7, 8, 9, 16, 17, 18, 19, 20 \}$ can be used to create the overlapping areas (Fig.~\ref{fig:flood_fill_3}).
However, these separator images are insufficient to compute the similarity transformations for merging all local reconstructions due to the sparsity of the view graph. 2) Graph cut tends to separate partitions along edges with weak associations. This means the separators are often weakly constrained during reconstruction and thus their poses might not be 
accurate enough during reconstruction.

We propose a flood-fill graph partition algorithm to overcome the above-mentioned disadvantages. We refer to the added nodes in each cluster after an expansion operation as a \textit{layer}. The separators are collected to form a layer after the graph cut on the complete view graph. Fig.~\ref{fig:flood_fill_2} shows examples of the separators marked green. We have separators $\mathcal{S}_1=\{\{3, 7, 8\}, \{9, 16, 17\}, \{18, 19, 20\}\}$ in the first layer. We then collect all the adjacent images of every separator for each partition. We find one adjacent image that does not belong to partition $k$, and add it to the second layer of separators $\mathcal{S}_2$ in partition $k$. Adjacent images are sorted in descending order according to the weights of the edges, i.e. the number of inlier matches.
Fig.~\ref{fig:flood_fill_3} shows that the separators $\mathcal{S}_2 = \{\{9, 20\}, \{8, 18\}, \{8, 16\}\}$ at the second layer
after traversing all separators in $\mathcal{S}_1$. The expansion step 
is repeated until the number of overlapping images reaches the overlapping threshold $\tau_{\text{ot}}$ (e.g. 30$\%$).Fig.~\ref{fig:flood_fill_4} shows the separators $\mathcal{S}_3$ at the third layer.
% The details of our flood-fill graph partition algorithm are given by Alg.~\ref{alg:adaptive_graph_partition}.
% \begin{algorithm}
%   \caption{Adaptive Flood-Fill Graph Partition Algorithm}
%   \label{alg:adaptive_graph_partition}
%   \begin{algorithmic}[1]
%     \Require Initial view graph $\mathcal{G}=\{\mathcal{V}, \mathcal{E}\}$,
%              Overlapping threshold $\tau_{\text{ot}}$, Partition number $K$
%     \Ensure Sub-graphs $\{ \mathcal{G}_k=\{\mathcal{V}_i, \mathcal{E}_i \} \mid i \in [0, K] \}$

%     \State Overlapping ratio $\tau_{or} := 0$, Separators $\mathcal{V}^{s} := \emptyset$, 
%            $\{ \mathcal{G}_k \} := \text{GraphCut}(\mathcal{G})$.

%     \While{$\tau_{\text{or}} < \tau_{\text{ot}}$}
%       \State Update separators $\mathcal{V}^{s} = \{\mathcal{V}^{s}_{0}, \cdots, \mathcal{V}^{s}_{K}\}$.
%       \State Edges $\mathcal{E}^{\text{dis}} := \emptyset$.
%       \For{$k \in [0, K]$}
%         \State $\mathcal{E}^{\text{dis}}_k = \mathcal{E} - \mathcal{E}_k$ and
%                $\mathcal{E}^{\text{dis}}_k$ contains $\mathcal{V}^{s}_{k}$.
%         \State $\mathcal{E}^{\text{dis}} += \mathcal{E}^{\text{dis}}_k$.
%       \EndFor

%       \State Sort $\mathcal{E}^{\text{dis}}$ by descending order.
%       \For{Edge $e \in \mathcal{E}^{\text{dis}}$}
%         \State Select a partition $\mathcal{G}_k$ contains one of the nodes in $e$ and has the smallest size.
%         \State Add $e$ to $\mathcal{G}_k$.
%       \EndFor

%       \State Update $\tau_{\text{or}}$.
%     \EndWhile

%   \end{algorithmic}
% \end{algorithm}
% % \vspace{-1cm}

\subsubsection{Local Incremental SfM}
\label{subsubsec:local_incremental_sfm}
We perform incremental SfM in parallel after graph partitioning. For local incremental SfM, we utilize the result of global SfM $\hat{\mathcal{P}}_{\text{global}}$ 
to improve the robustness of the image registration step, and to further constrain the camera poses during global optimization.

\paragraph{Image Registration}
\label{para:image_registration}

We follow~\cite{DBLP:conf/cvpr/SchonbergerF16} for the two-view initialization. 
We then select a batch of the next-best images to register, where any image that sees 
at least $v_p$ scene points are put into one batch and sorted in descending order. For each candidate image $i$, 
we first use the P3P~\cite{DBLP:conf/cvpr/KneipSS11} to compute the initial pose $\mathbf{P}_{i}^{\text{p3p}}$. 
However, images can be registered wrongly due to wrong matches or scene degeneration. We propose to 
also compute the image pose $\mathbf{P}_{i}^{\text{gb}}=[\mathbf{R}_{i}^{\text{gb}} \mid \mathbf{t}_{i}^{\text{gb}}]$ 
using $\hat{\mathcal{P}}_{\text{global}}$. We first collect the set of registered images that are co-visible to image $i$, 
and then the rotation of image $i$ can be computed by a single rotation averaging~\cite{DBLP:journals/ijcv/HartleyTDL13}:
\begin{small}
\vspace{-0.08in}
\begin{align}
  \argmin_{\mathbf{R}_{i}^{\text{gb}}} 
    & \sum_{k} \| \log (\hat{\mathbf{R}}_{ki} \mathbf{R}_k, \mathbf{R}_{i}^{\text{gb}}) \|, \quad \text{where}
    \quad \hat{\mathbf{R}}_{ki} = \hat{\mathbf{R}}_i \hat{\mathbf{R}}_k^\top,
\end{align}
\end{small}
where $k$ is the index of images that are co-visible to image $i$. 
For image translation, we first compute the translation of image $i$ by each co-visible image and
simply adopt the median of each dimension in translations $\mathbf{t}_{i}^{\text{gb}}$:
\begin{small}
\vspace{-0.1in}
\begin{align}
  & \mathbf{t}_{i}^{\text{gb}} = \text{median} 
    \{ \hat{\mathbf{t}}_{ki} + \hat{\mathbf{R}}_{ki} \mathbf{t}_k \}, \quad \text{where}
    \quad \hat{\mathbf{t}}_{ki} = \hat{\mathbf{t}}_i - \hat{\mathbf{R}}_{ki} \hat{\mathbf{t}}_k.
\end{align}
\end{small}
To select the best initial pose, we reproject all visible 3D points of image $i$ to compute the 
reprojection errors and mark the 3D point with the reprojection error less than 8px as an inlier. Finally, we select the pose which has the most inliers.

\begin{figure*}[htbp]
        \centering
        \includegraphics[width=1.0\linewidth]{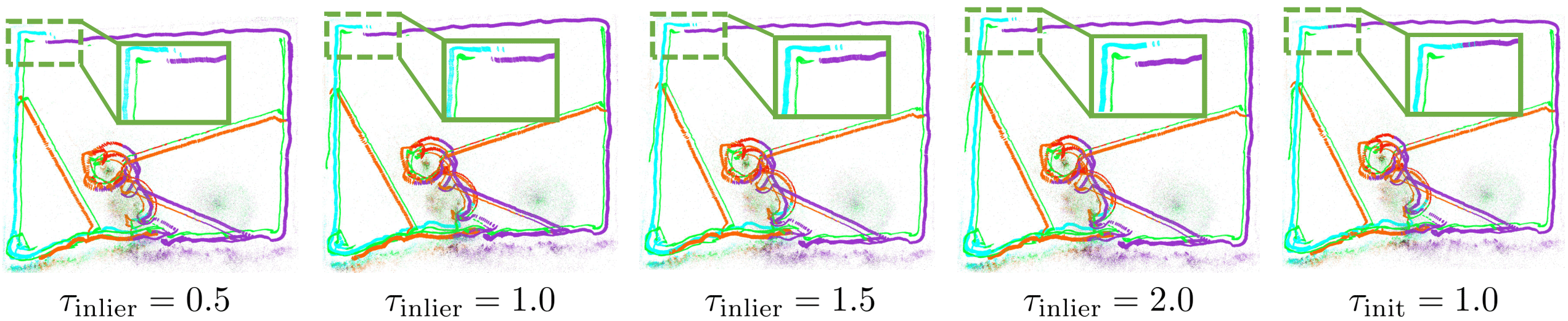}
        \vspace{-0.3in}

        \caption{\textbf{Vehicle trajectories of different threshold trials when merging sub-reconstructions}. The last figure 
                 is obtained by our method which starts from an 
                 initial inlier threshold $\tau_{\text{init}}$. Others are the results of using 
                 a fixed threshold during the alignment to merge all local reconstructions.}
        \label{fig:merge_trials}
        \vspace{-0.1in}
\end{figure*}

\paragraph{Bundle Adjustment}
\label{para:bundle_adjustment}

To alleviate the drift problem for local incremental SfM, we perform global optimization using the classical bundle adjustment with the absolute poses obtained from global SfM as the supervision for the incrementally registered poses, i.e.
% \begin{small}
% \begin{align}
%   \argmin_{ \mathbf{R}, \mathbf{C}, \mathbf{X}} 
%        & \Big\{ \sum_i \sum_k \| \mathbf{\Pi} (\mathbf{R}_i, \mathbf{C}_i, \mathbf{X}_k) - \mathbf{u}_{ik} \| ~~~+ \\ \nonumber
%        & \sum_{(i, j) \in \mathcal{E}_{\text{aug}}} \| \log (\mathbf{R}_j \mathbf{R}_i^\top, \hat{\mathbf{R}}_j\hat{\mathbf{R}}_i^\top) \| ~~~+ \\ \nonumber
%        & \sum_{(i, j) \in \mathcal{E}_{\text{aug}}} d_{\angle} (\mathbf{C}_i - \mathbf{C}_j, \hat{\mathbf{C}}_i - \hat{\mathbf{C}}_j) \Big\},
% \end{align}
% \end{small}
\begin{small}
  \begin{align}
    \argmin_{ \mathbf{R}, \mathbf{C}, \mathbf{X}} 
         & \Big\{ \sum_i \sum_k \| \mathbf{\Pi} (\mathbf{R}_i, \mathbf{C}_i, \mathbf{X}_k) - \mathbf{u}_{ik} \| ~~~+ \\ \nonumber
         & \sum_{(i, j) \in \mathcal{E}_{\text{aug}}} \Big ( \| \log (\mathbf{R}_{ij}, \hat{\mathbf{R}}_{ij} \| + 
                                                      d_{\angle} (\mathbf{t}_{ij}, \hat{\mathbf{t}}_{ij}) \Big ) \Big\},
  \end{align}
\end{small}
where $\mathbf{\Pi} (\cdot)$ reprojects a 3D point back to the image plane, $d_{\angle} (\cdot)$ 
denotes the angle between two vectors. Note that we do not make the 
hard constraint to force the translation part of $\hat{\mathbf{P}}_{ij}^{-1} \mathbf{P}_{ij}$ to be a zero-vector.
Instead, we use $d_{\angle} (\mathbf{t}_{ij}, \hat{\mathbf{t}}_{ij}) = d_{\angle} (\mathbf{C}_i - \mathbf{C}_j, \hat{\mathbf{C}}_i - \hat{\mathbf{C}}_j)$ 
to constrain the translation direction of camera poses. This is because the absolute positions obtained from global SfM are not sufficiently accurate.

% \vspace{-0.45in}
\subsubsection{Adaptive Global Alignment}
\label{subsubsec:adaptive_global_alignment}

The global alignment step is crucial for the divide-and-conquer SfM since a wrong similarity transformation can cause catastrophic failure of the reconstruction. The difficulties in estimating 
a reliable similarity transformation are due to 1) The existence of outliers in registered camera
poses. Although the outliers can be identified by RANSAC~\cite{DBLP:journals/cacm/FischlerB81}, the threshold that indicates outliers is hard to determine. This is due to the loss of the absolute
scale of the real world in SfM without additional information such as GPS. It indicates that \textit{the optimal outlier threshold varies for each cluster}. 2) The estimated similarity transformation can 
overfit wrongly with insufficient sample points. %, and thus it is not the desired similarity transformation. 
Existing divide-and-conquer methods
~\cite{DBLP:conf/accv/BhowmickPCGB14,Zhu2017Parallel,DBLP:conf/cvpr/ZhuZZSFTQ18,DBLP:conf/bmvc/FangPQ19,DBLP:journals/pr/ChenSCW20} 
suffer from the two issues because the similarity transformations can only be estimated from the overlapping areas between the pairwise local partitions.

To tackle the first issue, we propose an adaptive strategy to determine the inlier threshold 
$\tau_{\text{inlier}}$. Given an initial inlier threshold $\tau_{\text{init}}$, we first estimate the similarity transformation by 
RANSAC~\cite{DBLP:journals/cacm/FischlerB81}. We then compute the inlier ratio $r_{\text{inlier}}$ and increase the inlier threshold if $r_{\text{inlier}} < r_{\min}$. 
Furthermore, we decrease the threshold if $r_{\text{inlier}} \geq r_{\max}$ to prevent the threshold from becoming too large. A large threshold allows more outliers to be 
falsely selected and thus harming the similarity transformation estimation. The second issue can be solved easily within our framework. We set the coordinate frame of 
the global SfM as the reference frame, and align each local SfM into the reference frame. Therefore, for each partition, 
we can have as many sample points as the number of common registered images between a global SfM and a local partition to compute the similarity transformation. 
% The details of our global alignment algorithm are given by Alg.~\ref{alg:adaptive_global_alignment}. 
We also show the effectiveness of the algorithm to merge local reconstructions in Fig.~\ref{fig:merge_trials}. When zooming in, we can observe that our adaptive strategy perfectly 
closed the loop while other fixed threshold trials failed.

% \begin{figure*}[htbp]
%   \centering
%     \includegraphics[width=1.0\linewidth]{img/yht_merge_trials}

%   \caption{\textbf{Different threshold trials for merging sub-reconstructions}. The last figure 
%            is the result from Alg.~\ref{alg:adaptive_global_alignment} and starts from an 
%            initial inlier threshold $\tau_{\text{init}}$. Other figures are the results of using 
%            a fixed threshold during the alignment to merge all local reconstructions.}
%   \label{fig:merge_trials}
%   % \vspace{-0.3in}
% \end{figure*}
% \vspace{-1cm}

% % \vspace{-1cm}

%---------------------------------------------------------------------------------------------

\section{Experimental Results}
\label{sec:experiment}

In this section, we perform extensive experiments to demonstrate the accuracy, efficiency, 
and robustness of our proposed methods.

% \begin{figure*}[htbp]
%         \centering
%         \subfigure[HFNet~\cite{DBLP:conf/cvpr/SarlinCSD19}+NN]
%         {
%           \begin{minipage}{0.47\linewidth}
%             \label{fig:B6_hfnet_refine}
%             \centering
%             \includegraphics[width=1.0\linewidth]{img/B6_hfnet_refine}
%           \end{minipage}
%         }\quad
%         \subfigure[Superpoint~\cite{DBLP:conf/cvpr/DeToneMR18}+Superglue~\cite{DBLP:conf/cvpr/SarlinDMR20}]
%         {
%           \begin{minipage}{0.47\linewidth}
%             \label{fig:B6_superglue_refine}
%             \centering
%             \includegraphics[width=1.0\linewidth]{img/B6_superglue_refine}
%           \end{minipage}
%         }
%         \subfigure[Filtered matching pairs.]
%         {
%           \begin{minipage}{0.98\linewidth}
%             \label{fig:B6_match_pairs}
%             \centering
%             \includegraphics[width=1.0\linewidth]{img/B6_match_pairs}
%           \end{minipage}
%         }
%         \caption{\textbf{Vehicle trajectories after match refinement on B6 dataset}. In Fig.~\ref{fig:B6_hfnet_refine} and 
%                  Fig.~\ref{fig:B6_superglue_refine}, the visual results are respectively reconstructed 
%                  without (left) and with (right) match refinement in each sub-figure. Fig.~\ref{fig:B6_match_pairs} 
%                  shows the wrong matching pairs that are filtered by our method.}
%         \label{fig:match_refine_B6}
%         \vspace{-0.1in}
% \end{figure*}

\begin{figure*}[htbp]
        \centering
        \includegraphics[width=1.0\linewidth]{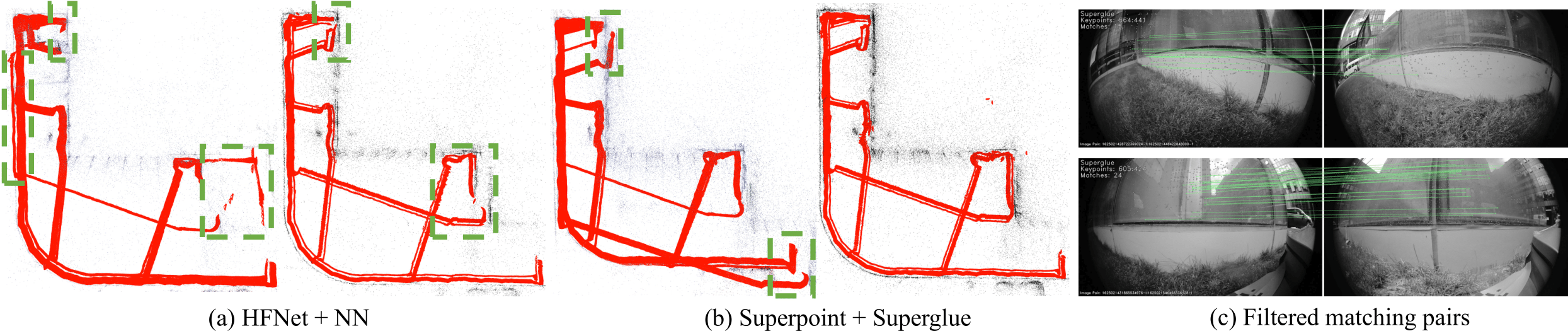}
        \vspace{-0.3in}

        \caption{\textbf{Vehicle trajectories after match refinement on B6 dataset}. In Fig.(a) and 
                 Fig.(b), the visual results are respectively reconstructed 
                 without (left) and with (right) match refinement in each sub-figure. Fig.(c) 
                 shows some of the wrong matching pairs that are filtered by our method.}
        \label{fig:match_refine_B6}
        \vspace{-0.1in}
\end{figure*}

% ------------------------------------------ Table ---------------------------------------------------
\begin{table*}[htbp]
        \centering
        \resizebox{1.0\textwidth}{!}{
          \begin{tabular}{c c r r r r r r r r r r r r r r r r r r r}
            \toprule
      
            \multirow{2}{*}{Dataset}  &
            \multirow{2}{*}{$N$} &
              \multicolumn{5}{c}{\textbf{COLMAP}~\cite{DBLP:conf/cvpr/SchonbergerF16}} &
              \multicolumn{5}{c}{\textbf{GraphSfM}~\cite{DBLP:journals/pr/ChenSCW20}} &  
              \multicolumn{4}{c}{\textbf{Ours(Global SfM)}} & 
              \multicolumn{5}{c}{\textbf{Ours(Global+Inc.)}} \\
              \cmidrule(r){3-7} \cmidrule(r){8-12} \cmidrule(r){13-16} \cmidrule{17-21} 
                  & \ & \makecell[c]{$N_c$} & \makecell[c]{$N_p$} & \makecell[c]{$\bar{L}$} & \makecell[c]{$ \text{RMSE} $} & \makecell[c]{$T$}
                      & \makecell[c]{$N_c$} & \makecell[c]{$N_p$} & \makecell[c]{$\bar{L}$} & \makecell[c]{$ \text{RMSE} $} & \makecell[c]{$T$}
                      & \makecell[c]{$N_c$} & \makecell[c]{$N_p$} & \makecell[c]{$\bar{L}$} & \makecell[c]{$T$} %& $ \text{RMSE} $ & $T$
                      & \makecell[c]{$N_c$} & \makecell[c]{$N_p$} & \makecell[c]{$\bar{L}$} & \makecell[c]{$ \text{RMSE} $} & \makecell[c]{$T$} \\
            \midrule
      
              high free & 48,753
                        & \textcolor{red}{48,733} & 567,030 & 21.59 & 1.47 & \textcolor{MyBlue}{597,171}
                        & \textcolor{MyBlue}{48,491} & 540,711 & 22.73 & 1.38 & \textcolor{red}{88,896} ($\times 6.7 \uparrow $)
                        & 48,758 & 521,080 & 14.51 & 5,177 % & 1.08 & 5,177
                        & \textcolor{MyGreen}{48,694} & 540,942 & 22.79 & 1.66 & \textcolor{MyGreen}{105,163} ($\times 5.7 \uparrow $) \\
      
              711 & 29,619 
                  & \textcolor{MyGreen}{27,175} & 303,352  & 25.35 & 1.64 & \textcolor{MyBlue}{160,322}
                  & \textcolor{MyBlue}{29,618} & 259,292 & 33.37 & 1.46 & \textcolor{red}{33,514} ($\times 4.8 \uparrow $)
                  & 29,629 & 249,673 & 18.86 & 3,499 % & 1.21 & 3,499
                  & \textcolor{red}{29,619} & 256,495 & 33.79 & 1.61 & \textcolor{MyGreen}{38,682} ($\times 4.1 \uparrow $) \\
      
              yht & 7,472 
                  & \textcolor{MyGreen}{7,470} & 90,437  & 20.81 & 1.16 & \textcolor{MyBlue}{20,428}
                  & \textcolor{MyBlue}{6,709} & 78,659 & 20.58 & 1.17 & \textcolor{red}{7,526} ($\times 2.7 \uparrow $)
                  & 7,472 & 132,167 & 13.67 & 524 % & 1.41 & 524
                  & \textcolor{red}{7,472} & 108,711 & 17.35 & 1.43 & \textcolor{MyGreen}{9,778} ($\times 2.1 \uparrow $) \\
      
              A4 & 5,184 
                 & \textcolor{MyGreen}{5,132} & 33,694  & 41.92 & 1.69 & \textcolor{MyBlue}{18,104}
                 & \textcolor{MyBlue}{4,285} & 28,726 & 49.79 & 1.55 & \textcolor{MyGreen}{12,670} ($\times 1.4 \uparrow $)
                 & 5,184 & 24,193 & 26.59 & 1,349 % & 1.21 & 1,349
                 & \textcolor{red}{5,184} & 34,007 & 48.30 & 1.43 & \textcolor{red}{6,924} ($\times 2.6 \uparrow $) \\
      
              Htbd & 14,651 
                 & \textcolor{MyGreen}{14,645} & 231,870  & 24.62 & 1.30 & \textcolor{MyBlue}{56,888}
                 & \textcolor{MyGreen}{14,645} & 232,441 & 24.25 & 1.37 & \textcolor{MyGreen}{17,187} ($\times 3.3 \uparrow $)
                 & 14,646 & 190,904 & 23.47 & 1,523 % & 1.07 & 1,523
                 & \textcolor{red}{14,646} & 238,035 & 23.76 & 1.36 & \textcolor{red}{16,852} ($\times 3.4 \uparrow $) \\
      
              jy1 & 32,484 
                 & \textcolor{MyBlue}{32,463} & 534,117 & 20.57 & 1.44 & \textcolor{MyBlue}{346,161}
                 & \textcolor{red}{32,466} & 536,331 & 20.18 & 1.52 & \textcolor{red}{28,673} ($\times 12.1 \uparrow $)
                 & 32,484 & 463052 & 16.12 & 3,077 % & 1.07 & 3,077
                 & \textcolor{red}{32,466} & 621,437 & 17.77 & 1.53 & \textcolor{MyGreen}{33,555} ($\times 10.3 \uparrow $) \\
            \bottomrule
          \end{tabular}
        }
      
        \caption{Comparison of runtime and accuracy on real-world datasets. For runtime $T$ (seconds), 
          the \textcolor{red}{first}, \textcolor{MyGreen}{second} and \textcolor{MyBlue}{third} 
          The best results are highlighted in color. $N_c, N_p$ denote the number of registered 
          images and 3D points, respectively, $\bar{L}$ denotes the average track length
          , and $\text{RMSE}$ denotes the root mean square error in pixel.}
        \label{table:real_world_data}
        \vspace{-0.45in}
\end{table*}

\vspace{-0.12in}
\subsection{Implementation Details}
\label{subsec:implementation_details}

We use HFNet~\cite{DBLP:conf/cvpr/SarlinCSD19} as the default feature extractor and use 
the NN search for matching. A maximum of 500 feature points are extracted from each image and 
matched to the top 30 most similar images based on the global descriptors from HFNet. We assume cameras are pre-calibrated and 
use the ceres-solver~\cite{ceres-solver} for bundle adjustment. %Our method and GraphSfM~\cite{DBLP:journals/pr/ChenSCW20} are evaluated on a single computer in parallel. 
We did not compare our method against~\cite{DBLP:conf/icit2/JiangTMO21}, 
as VINs~\cite{DBLP:journals/trob/QinLS18} fails to find the right loops in our datasets.
All methods are run on the same computer with 40 CPU cores and 96 GB RAM.

\textbf{Evaluation Datasets:} We evaluate our method on our self-collected outdoor datasets and the 4seasons~\cite{DBLP:conf/dagm/WenzelWYCKSZC20} 
datasets. Our self-collected datasets are collected by low-speed autonomous mowers, of which the running environments have many plants and 
texture-less areas. %Besides, each mower is mounted with one fisheye camera, which has only a limited field of view.
The 4seasons dataset is a 
cross-season dataset that includes multi-sensor data such as IMU, GNSS, and stereo images. It also provides camera poses computed by 
VI-Stereo-DSO~\cite{DBLP:conf/iccv/WangSC17,DBLP:conf/icra/StumbergUC18} and ground-truth camera poses by fusing multi-sensor data into a 
SLAM system. See our attached video for a more qualitative and quantitative evaluation of the 4Seasons dataset.

\textbf{Running Parameters:} Empirically, we use the time threshold $\epsilon_{T} = 500~\text{ms}$ to adopt the fused relative poses 
in $\mathcal{G}_{\text{aug}}$, and $\epsilon_{\mathbf{R}} = 5~\text{degree}$ to check to relative rotation discrepancy. 
The point-to-epipolar line distance is $\epsilon_{M} = 4~\text{px}$. Besides, we set the overlapping ratio 
$\tau_{\text{ot}}=0.3$ in the graph partition, $v_p=10$ for an image to be a candidate to register, 
and $r_{\min}=0.7, r_{\max}=0.9, \tau_{\text{init}}=1.0, \alpha_{\text{inc}}=0.2, \alpha_{\text{dec}}=0.1$ in global alignment.

\vspace{-1mm}
\subsection{How Matching Refinement Saves SfM?}
\label{subsec:how_matching_refinement_saves_sfm}
\vspace{-1mm}

In addition to running our experiments on HFNet, we also do evaluations on different trials.
We first show the reconstruction results conducted on a challenging scene in Fig.~\ref{fig:match_refine_B6}, which is
difficult for visual methods to identify the wrong feature matches due to specular issues.
% two different scenes in 
% Fig.~\ref{fig:match_refine_B6} and Fig.~\ref{fig:match_refine_711}. 
% Specifically, Fig.~\ref{fig:B6_match_pairs} is a scene with mirrors 
% and Fig.~\ref{fig:711_match_pairs} is a scene with many similar structures. 
% Both scenes are difficult for visual methods to identify the wrong feature matches.

We use two different combinations of methods for feature extraction and matching in each scene. In the first combination, 
we use HFNet~\cite{DBLP:conf/cvpr/SarlinCSD19} for feature extraction and NN search for feature matching. In the second combination, 
we use Superpoint~\cite{DBLP:conf/cvpr/DeToneMR18} for feature extraction and Superglue~\cite{DBLP:conf/cvpr/SarlinDMR20} for feature matching. Both settings use RANSAC
~\cite{DBLP:journals/cacm/FischlerB81} to remove matching outliers that do not satisfy the point-to-epipolar line constraint. In each sub-figure, the left and right 
images are the results without and with matching refinement, respectively. It can be seen that for HFNet + NN, while both methods fail to reconstruct the two datasets, the result after
our result is visually better than without matches refinement.
For Superpoint + Superglue, the state-of-the-art methods respectively on feature extraction and matching, also fails on the dataset without refining matches. 
In contrast, our method can correctly identify the wrong matching pairs and then leverage the refined matchings to greatly improve the reconstruction quality for both settings.

% \begin{figure*}[htbp]
%   \centering
%   \subfigure[HFNet~\cite{DBLP:conf/cvpr/SarlinCSD19}+NN]
%   {
%     \begin{minipage}{0.47\linewidth}
%       \label{fig:711_hfnet_refine}
%       \centering
%       \includegraphics[width=1.0\linewidth]{img/711_hfnet_refine}
%     \end{minipage}
%   }\quad
%   \subfigure[Superpoint~\cite{DBLP:conf/cvpr/DeToneMR18}+Superglue~\cite{DBLP:conf/cvpr/SarlinDMR20}]
%   {
%     \begin{minipage}{0.47\linewidth}
%       \label{fig:711_superglue_refine}
%       \centering
%       \includegraphics[width=1.0\linewidth]{img/711_superglue_refine}
%     \end{minipage}
%   }
%   \subfigure[Filtered matching pairs.]
%   {
%     \begin{minipage}{0.98\linewidth}
%       \label{fig:711_match_pairs}
%       \centering
%       \includegraphics[width=1.0\linewidth]{img/711_match_pairs}
%     \end{minipage}
%   }
% %   \vspace{-0.2in}
%   \caption{\textbf{Vehicle trajectories after match refinement on 711 dataset}. Fig.~\ref{fig:711_hfnet_refine} and 
%           Fig.~\ref{fig:711_superglue_refine} show the reconstructions 
%           without (left) and with (right) match refinement. Fig.~\ref{fig:711_match_pairs} 
%           shows the wrong matching pairs that are filtered out by our method.}
%   \label{fig:match_refine_711}
%   % \vspace{-0.2in}
% \end{figure*}

\vspace{-2mm}
\subsection{Qualitative Evaluation on Real-World Datasets}
\label{subsec:qualitative_evaluation_on_real_world_datasets}
\vspace{-1mm}

We evaluated our full pipeline on several outdoor datasets. We use the registered images number $N_c$, the recovered 3D 
points $N_p$, the average track length $\bar{L}$, and the root mean square error (RMSE) to evaluate the 
qualitative accuracy. As shown in Table.~\ref{table:real_world_data}, our method shows the 
most number of registered images in almost all the datasets, while \cite{DBLP:journals/pr/ChenSCW20} shows the least number of registered images.
In terms of efficiency, our method is moderately slower than 
GraphSfM~\cite{DBLP:journals/pr/ChenSCW20} in most datasets since our method requires an additional global 
SfM reconstruction step. Interestingly, GraphSfM~\cite{DBLP:journals/pr/ChenSCW20} is almost $1\times$ slower 
than our method on the A4 dataset. We conjecture that it is due to the frequent failure of GraphSfM in 
selecting suitable images to register and therefore more trials are required to register as many images 
as possible. On the other hand, our method is robust enough to deal with the case since we get the initial 
poses of the images from P3P or global SfM. Our explanation is validated in Table.~\ref{table:real_world_data} 
where GraphSfM~\cite{DBLP:journals/pr/ChenSCW20} recovers only 4,235 poses out of 5,184 images, which is almost 20\% less than our method.
We can further notice that the average track length of global SfM is remarkably shorter than other methods, which 
means poses from global SfM are not accurate. %The teaser figure also shows the reconstruction result of the high free dataset.

%----------------------------------------------------------------------
\vspace{-2mm}
\section{Conclusion}
\label{sec:conclusion}

In this paper, we proposed a robust SfM method that is adaptive to scenes in different scales and 
environments. Integrating data from low-cost sensors, our initial global SfM can benefit from the augmented view graph, 
where the solvability of the original view graph is enhanced. The global SfM result is used as a reliable pose prior to improve the 
robustness of the subsequent local incremental SfM and the final global alignment steps. Comprehensive 
experiments on different challenging scenes demonstrated the robustness and adaptivity of our method, 
whilst taking more computation burden with an additional global SfM step.

% \noindent \textbf{Limitations and Future Work: } Our framework relies on a coarse global SfM. However, when the 
% scale of datasets grows larger (e.g. millions of images), the time complexity of the global SfM is also non-negligible. 
% The distributed motion averaging method~\cite{DBLP:conf/cvpr/ZhuZZSFTQ18} can be adopted,
% but the solvability of the sub-graphs might be reduced due to the loss of constraints after partitioning.
% In our future work, we plan to analyze the solvability of the multi-graphs motion 
% synchronization~\cite{Cin_2021_ICCV} and solve the distributed multi-graphs-based global SfM robustly.

\noindent \textbf{Acknowledgement.} This research/project is supported by the National Research Foundation, Singapore under 
its AI Singapore Programme (AISG Award No: AISG2-RP-2021-024), and the Tier 2 grant MOE-T2EP20120-0011 from the Singapore 
Ministry of Education.

% \clearpage
\newpage

{\small
\bibliographystyle{IEEEtran}
\bibliography{IEEEfull}
}

\clearpage
\newpage

\section{APPENDIX}

\subsection{Adaptive Flood-Fill Graph Partition Algorithm}

The pseudo-code of our adaptive flood-fill graph partition algorithm is given in Alg.~\ref{alg:adaptive_graph_partition}.

\begin{algorithm}
  \caption{Adaptive Flood-Fill Graph Partition Algorithm}
  \label{alg:adaptive_graph_partition}
  \begin{algorithmic}[1]
    \Require Initial view graph $\mathcal{G}=\{\mathcal{V}, \mathcal{E}\}$,
             Overlapping threshold $\tau_{\text{ot}}$, Partition number $K$
    \Ensure Sub-graphs $\{ \mathcal{G}_k=\{\mathcal{V}_i, \mathcal{E}_i \} \mid i \in [0, K] \}$

    \State Overlapping ratio $\tau_{or} := 0$, Separators $\mathcal{V}^{s} := \emptyset$, 
           $\{ \mathcal{G}_k \} := \text{GraphCut}(\mathcal{G})$.

    \While{$\tau_{\text{or}} < \tau_{\text{ot}}$}
      \State Update separators $\mathcal{V}^{s} = \{\mathcal{V}^{s}_{0}, \cdots, \mathcal{V}^{s}_{K}\}$.
      \State Edges $\mathcal{E}^{\text{dis}} := \emptyset$.
      \For{$k \in [0, K]$}
        \State $\mathcal{E}^{\text{dis}}_k = \mathcal{E} - \mathcal{E}_k$ and
               $\mathcal{E}^{\text{dis}}_k$ contains $\mathcal{V}^{s}_{k}$.
        \State $\mathcal{E}^{\text{dis}} += \mathcal{E}^{\text{dis}}_k$.
      \EndFor

      \State Sort $\mathcal{E}^{\text{dis}}$ by descending order.
      \For{Edge $e \in \mathcal{E}^{\text{dis}}$}
        \State Select a partition $\mathcal{G}_k$ contains one of the nodes in $e$ and has the smallest size.
        \State Add $e$ to $\mathcal{G}_k$.
      \EndFor

      \State Update $\tau_{\text{or}}$.
    \EndWhile

  \end{algorithmic}
\end{algorithm}

\subsection{Adaptive Global Alignment Algorithm}

The pseudo-code of our adaptive global alignment algorithm is given in Alg.~\ref{alg:adaptive_global_alignment}.

\begin{algorithm}
  \caption{Adaptive Global Alignment Algorithm}
  \label{alg:adaptive_global_alignment}
  \begin{algorithmic}[1]
    \Require Local reconstructions $\mathcal{M} = \{\mathbf{M}_i\}$, 
             $\tau_{\text{init}}, r_{\min}, r_{\max}, \text{iterNum}_{\max}, \alpha_{\text{inc}}, \alpha_{\text{dec}}$
    \Ensure Final reconstruction

    \For{$i < |\mathcal{M}|$}
      \State $\tau_{\text{inlier}}:=\tau_{\text{init}}, r_{\text{inlier}}:=0, \text{iterNum}:=0$

      \While{$r_{\text{inlier}} < r_{\min}$ \& $\text{iterNum} < \text{iterNum}_{\max}$}
        \State $\text{iterNum} := \text{iterNum} + 1$;
        \State Compute $\text{sim3}$ by $\tau_{\text{inlier}}$;
        \State Compute $r_{\text{inlier}}$ by $\text{sim3}$;
        
        \If{$r_{\text{inlier}} < r_{\min}$}
          \State $\tau_{\text{inlier}} := \tau_{\text{inlier}} + \alpha_{\text{inc}}$;
        \ElsIf{$r_{\text{inlier}} \geq r_{\max}$}
          \State $\tau_{\text{inlier}} := \tau_{\text{inlier}} - \alpha_{\text{dec}}$;
        \EndIf
      \EndWhile
    \EndFor
    
  \end{algorithmic}
\end{algorithm}

\subsection{Visualization Results on Self-Collected Dataset}

\begin{figure*}[htbp]
  \centering
  % \vspace{-0.4in}
  \subfigure[Qualititve comparison on the 711 dataset.]
  {
    \begin{minipage}{0.9\linewidth}
      \label{fig:711_20210705_compare}
      \centering
      \includegraphics[width=1.0\linewidth]{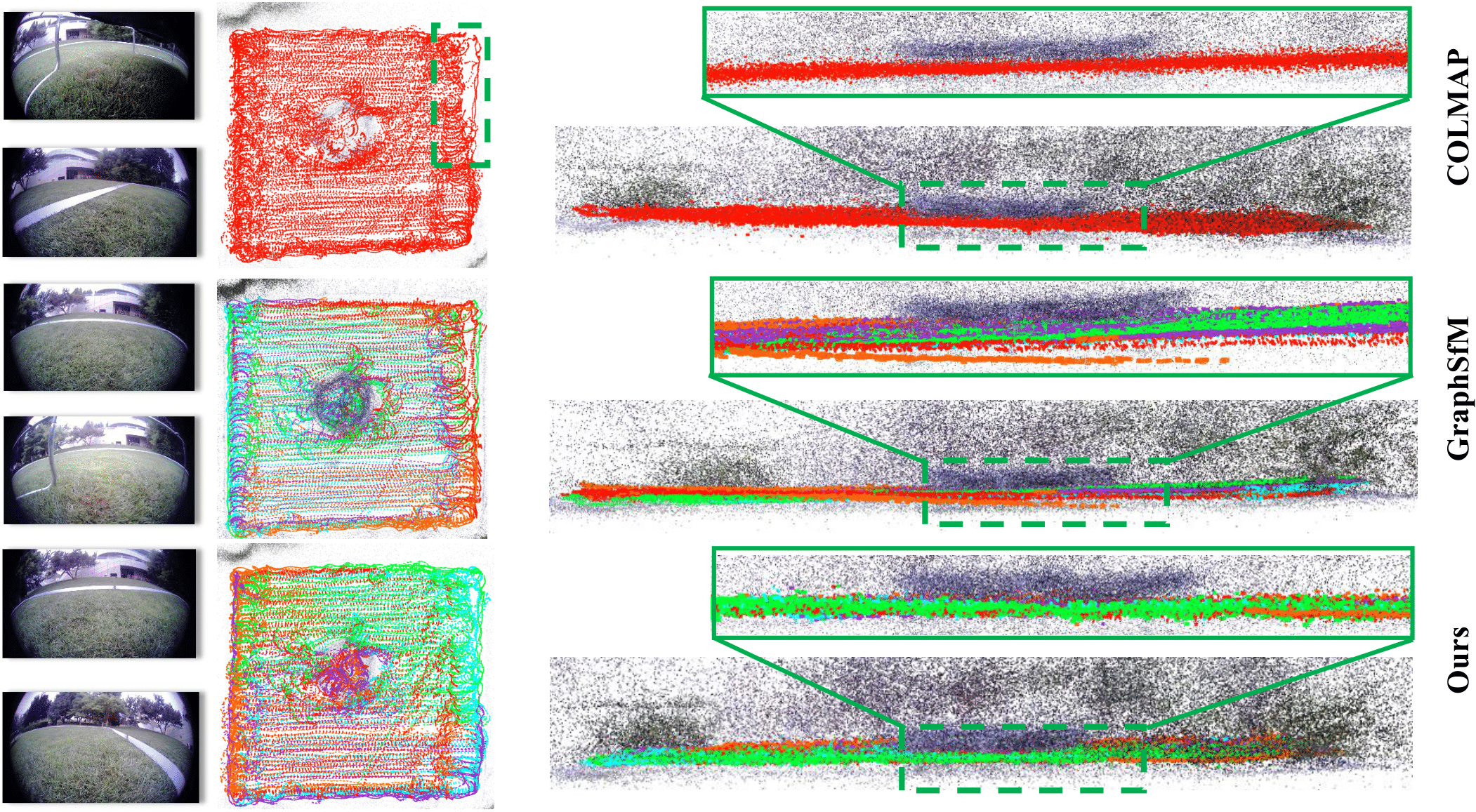}
    \end{minipage}
  } %\quad
  \subfigure[Qualititve comparison on the A4 dataset.]
  {
    \begin{minipage}{0.9\linewidth}
      \label{fig:A4_20210324_compare}
      \centering
      \includegraphics[width=1.0\linewidth]{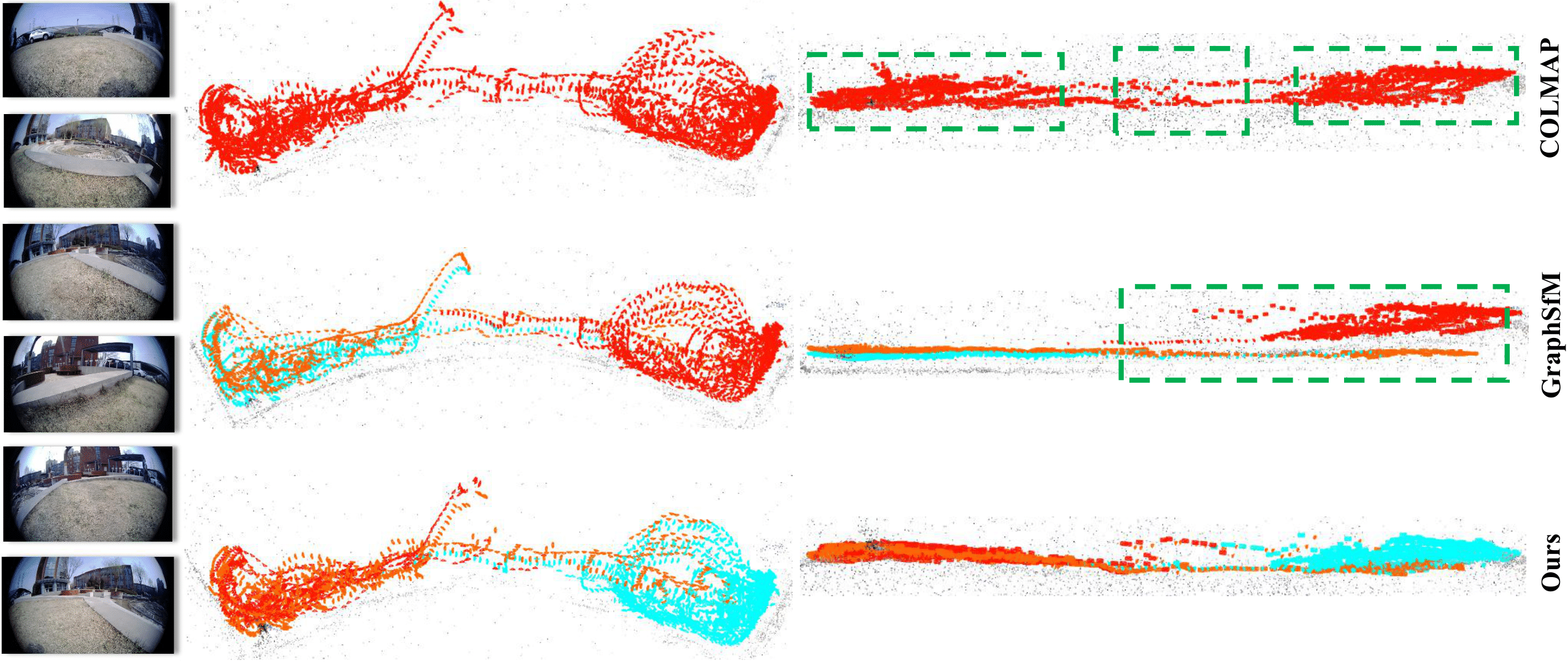}
    \end{minipage}
  }
  \quad %\vspace{-0.2in}
  \subfigure[Qualititve comparison on the high free dataset.]
  {
    \begin{minipage}{0.9\linewidth}
      \label{fig:high_free_20210519_compare}
      \includegraphics[width=1.0\linewidth]{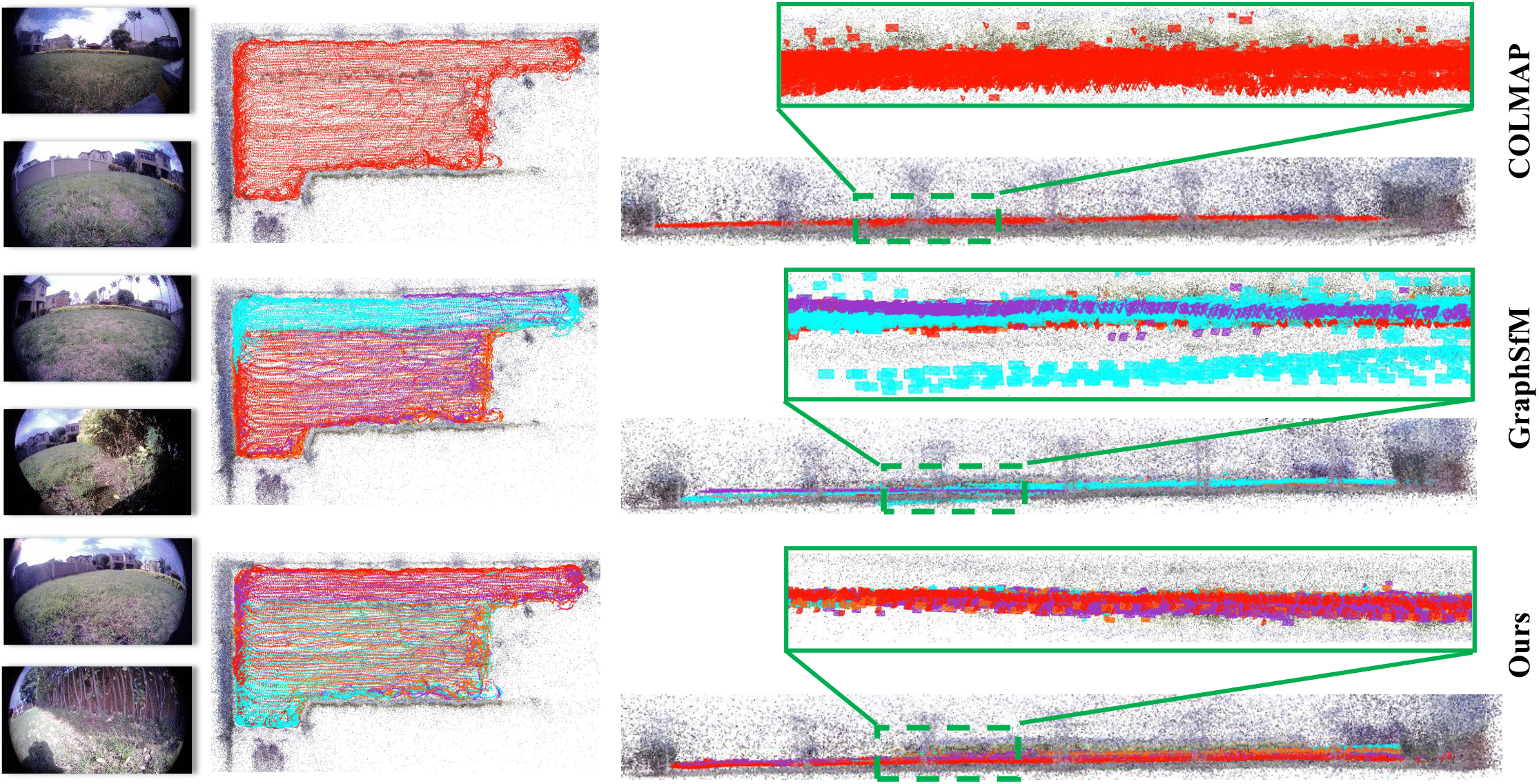}
    \end{minipage}
  }\quad

  \caption{Reconstruction comparisons on our self-collected dataset. From left to right are the input images, top-view reconstruction, and side-view reconstruction.}
  \label{fig:qualititve_recon_results}
\end{figure*}

The qualitative visualization results are shown in Fig.~\ref{fig:qualititve_recon_results}.
We can see that our reconstruction results are better than COLMAP~\cite{DBLP:conf/cvpr/SchonbergerF16} and 
GraphSfM~\cite{DBLP:journals/pr/ChenSCW20}, especially when we zoom in to 
see the image poses. Moreover, GraphSfM~\cite{DBLP:journals/pr/ChenSCW20} fails to correctly merge the 
sub-reconstructions. % in the 711, A4, and high free datasets.
The misalignment can be observed from the zoom-in areas of side-view images, which further validates the robustness of our method.

\subsection{Ablations of Augmented View Graph}

\begin{figure*}[ht]
  \centering

  \includegraphics[width=0.98\linewidth]{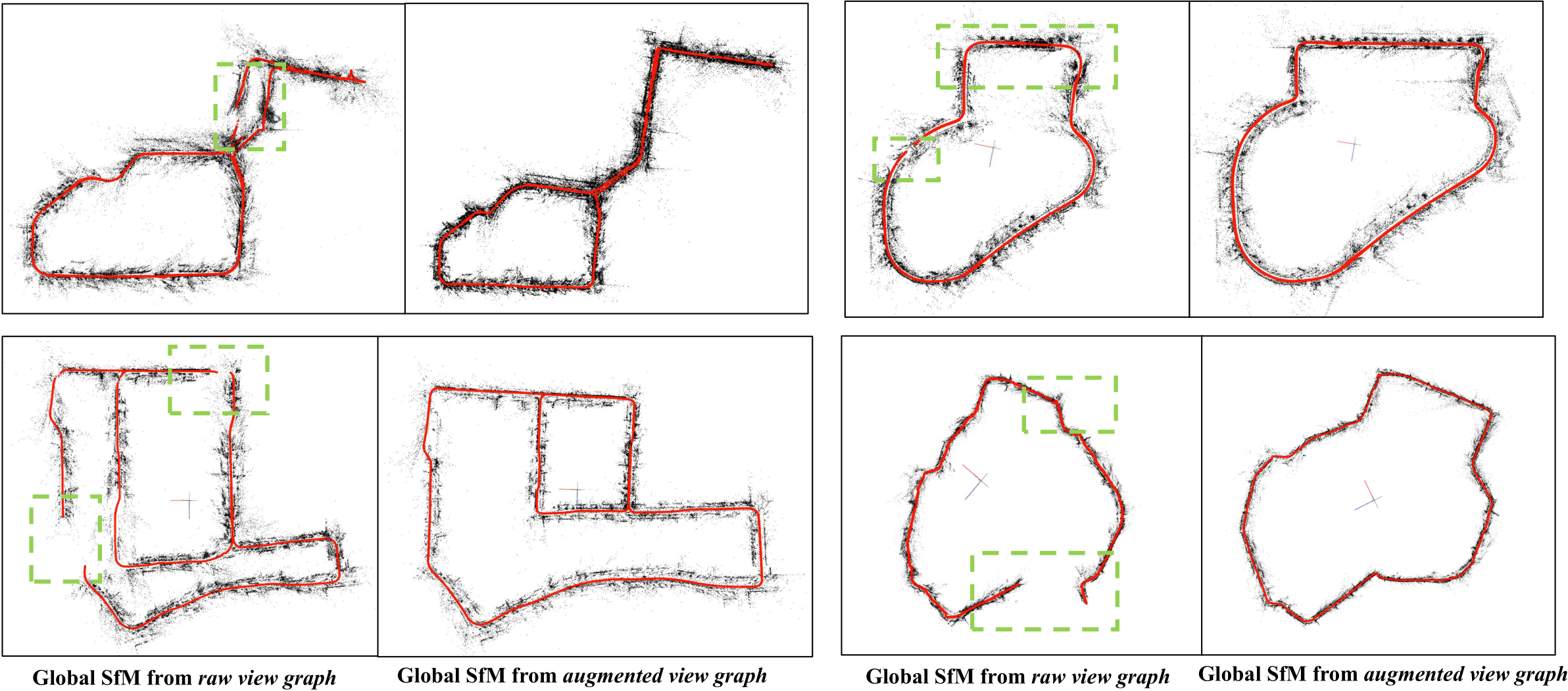}
  \caption{Ablations of our augmented view graph on the 4Seasons dataset.}
  \label{fig:ablation_aug_vg}
\end{figure*}

We present more ablation of the augmented view graph on the 4Seasons dataset in Fig.~\ref{fig:ablation_aug_vg}. More visualization results 
on this dataset can be seen in our attached video.

\subsection{Quantitative Results on 4Seasons dataset.}

\begin{table*}[htbp]
  \centering
  \resizebox{1.0\textwidth}{!}{
    \begin{tabular}{c c r r r r r r r r r r r r r r r}
      \toprule

      \multirow{2}{*}{\textbf{Scene}}  &
      \multirow{2}{*}{\textbf{Sequence}} &
      \multicolumn{5}{c}{\textbf{COLMAP}~\cite{DBLP:conf/cvpr/SchonbergerF16}} &
      \multicolumn{5}{c}{\textbf{Ours (Global SfM)}} &  
      \multicolumn{5}{c}{\textbf{Ours (final)}} \\
      \cmidrule(r){3-7} \cmidrule(r){8-12} \cmidrule(r){13-17}
      & \ & \makecell[c]{$N_c$} & \makecell[c]{$N_p$} & \makecell[c]{$\Delta \mathbf{R}$} & \makecell[c]{$\Delta \mathbf{t}$} & \makecell[c]{$T$}
          & \makecell[c]{$N_c$} & \makecell[c]{$N_p$} & \makecell[c]{$\Delta \mathbf{R}$} & \makecell[c]{$\Delta \mathbf{t}$} & \makecell[c]{$T$}
          & \makecell[c]{$N_c$} & \makecell[c]{$N_p$} & \makecell[c]{$\Delta \mathbf{R}$} & \makecell[c]{$\Delta \mathbf{t}$} & \makecell[c]{$T$} \\
      \midrule

      \multirow{7}{*}{Neighborhood}
        & recording\_2020-10-07\_14-53-52
        & 6,326 & 137,135 & \textbf{0.65} & 1.78 & 334.90
        & 6,036 &  66,777 & 2.52 & 1.17 &  14.68
        & 6,033 & 109,483 & 0.74 & \textbf{0.52} & 123.96 \\
      
        & recording\_2020-12-22\_11-54-24
        & 6,518 & 127,892 & 0.55 & 3.68 & 354.35
        & 6,144 &  64,405 & 1.10 & 0.86 &  15.83
        & 6,144 & 102,857 & \textbf{0.51} & \textbf{0.62} & 151.88 \\

        & recording\_2020-03-26\_13-32-55
        & 7,414 & 148,848 & \textbf{0.61} & 1.24 & 603.13
        & 5,982 &  70,066 & 0.92 & \textbf{0.79} &  17.10
        & 5,982 & 111,807 & 1.11 & 0.98 & 157.76 \\

        & recording\_2020-10-07\_14-47-51
        & 6,688 & 152,307 & \textbf{0.56} & 1.67 & 359.03
        & 6,248 &  76,305 & 2.20 & 1.17 &  15.70
        & 6,248 & 121,657 & 0.75 & \textbf{0.74} & 152.85 \\

        & recording\_2021-02-25\_13-25-15
        & 6,174 & 138,807 & 0.75 & 1.05 & 325.65
        & 5,238 &  62,879 & 1.00 & 1.14 &  15.12
        & 5,238 & 106,609 & \textbf{0.46} & \textbf{0.81} & 202.85 \\

        & recording\_2021-05-10\_18-02-12
        & 7,784 & 149,528 & 3.04 & 9.57 & 444.85
        & 5,834 &  61,889 & 1.49 & 1.38 &  12.76
        & 5,834 & 101,102 & \textbf{0.47} & \textbf{0.59} & 153.36 \\

        & recording\_2021-05-10\_18-32-32
        & 7,174 & 141,864 & 2.77 & 19.15 & 416.34
        & 6,046 &  89,010 & \textbf{1.14} & \textbf{1.03} & 23.81
        & 6,046 & 142,430 & 1.49 & 1.34 & 264.75 \\
      
      \hline

      \multirow{3}{*}{Business Park}
      & recording\_2021-01-07\_13-12-23
      & 8,016 & 109,399 & 0.72 & 0.75 & 643.22
      & 9,010 &  72,096 & 1.76 & 1.60 &  56.16
      & 9,010 & 100,057 & \textbf{0.66} & \textbf{0.51} & 465.34 \\
    
      & recording\_2020-10-08\_09-30-57
      & 11,520 & 127,013 & \textbf{0.37} & 1.57 & 1284.44
      & 8,278 &  66,087 & 1.59 & 1.51 &  48.72
      & 8,278 & 108,000 & 0.63 & \textbf{0.45} & 366.81 \\

      & recording\_2021-02-25\_14-16-43
      & 7,414 & 148,848 & \textbf{0.61} & 1.24 & 603.13
      & 5,982 &  70,066 & 0.92 & \textbf{0.79} &  17.10
      & 5,982 & 111,807 & 1.11 & 0.98 & 157.76 \\

      \hline

      \multirow{3}{*}{Old Town}
      & recording\_2020-10-08\_11-53-41
      & 19,332 & 279,989 & - & - & 2454
      & 12,910 & 181,569 & 2.23 & 2.81 &  45.72
      & 12,048 & 279,127 & \textbf{0.55} & \textbf{0.56} & 254.71 \\
    
      & recording\_2021-01-07\_10-49-45
      & 16.420 & 307,383 & 8.63 & 360.51 & 1496.6
      & 12,728 &  194,340 & 2.56 & 3.14 &  53.18
      & 12,728 & 327,348 & \textbf{1.55} & \textbf{1.03} & 238.82 \\

      & recording\_2021-02-25\_12-34-08
      & 18,950 & 305,461 & - & - & 2392.98
      & 12,387 & 182,940 & 2.02 & 3.14 &  40.97
      & 12,387 & 302,833 & \textbf{0.63} & \textbf{0.74} & 683.97 \\

      \hline

      \multirow{6}{*}{Office Loop}
      & recording\_2020-03-24\_17-36-22
      & 10,188 & 209,942 & 1.17 & 3.40 & 822.38
      & 9,522 &  126,680 & 2.28 & 2.38 &  31.87
      & 9,377 & 214,285 & \textbf{0.97} & \textbf{0.98} & 166.54 \\
    
      & recording\_2020-03-24\_17-45-31
      & 8,582 & 195,738 & 0.92 & 3.04 & 865.48
      & 9,186 &  122,713 & 2.79 & 2.20 &  33.91
      & 8,940 & 205,790 & \textbf{0.84} & \textbf{0.85} & 209.06 \\

      & recording\_2020-04-07\_10-20-31
      & 10,350 & 223.649 & 4.22 & 42.44 & 795.68
      & 10,184 &  138,446 & 2.53 & 1.78 &  39.83
      & 10,184 & 224,499 & \textbf{1.47} & \textbf{1.14} & 253.24 \\

      & recording\_2020-06-12\_10-10-57
      & 9,990 & 236,593 & 18.97 & 83.94 & 705.93
      & 10,150 & 164,062 & 1.92 & 1.61 & 37.32
      & 10,150 & 246,516 & \textbf{0.76} & \textbf{0.87} & 206.48 \\

      & recording\_2021-01-07\_12-04-03
      & 9,164 & 475,950 & \textbf{0.71} & 2.58 & 1000.75
      & 10,300 & 143,715 & 3.32 & 2.39 & 48.68
      & 10,300 & 223,676 & 1.08 & \textbf{0.67} & 249.42 \\

      & recording\_2021-02-25\_13-51-57
      & 9,574 & 214,695 & \textbf{0.84} & 2.84 & 773.32
      & 9,426 & 122,746 & 3.80 & 2.68 & 28.96
      & 9,426 & 204,289 & 1.01 & \textbf{0.91} & 173.29 \\

      \bottomrule
    \end{tabular}
  }

  \caption{Comparison of runtime and accuracy on the 4Seasons datasets. $T$ denotes the runtime (in minutes), 
    $N_c, N_p$ denote the number of registered images and 3D points, respectively, $\Delta \mathbf{R}, \Delta \mathbf{t}$ denotes 
    the mean rotation error (in degrees) and translation error (in meters), respectively, and we highlight the best results in bold.}
  \label{table:4season_quantitive_data}

\end{table*}

We present the quantitative results on the 4Seasons dataset in Table.~\ref{table:4season_quantitive_data}. The 4Seasons dataset provides ground 
truth camera poses and trajectories from VI-Stereo-DSO~\cite{DBLP:conf/iccv/WangSC17,DBLP:conf/icra/StumbergUC18}. The sensor data contain 
IMU, GNSS, and stereo images. In our experiment, we do not use the GNSS data. Besides, as this dataset does not provide wheel encode data, we 
perturb the VI-Stereo-DSO trajectories by Gaussian noise in the x-y-z axes to synthesize wheel encoder data. We strongly recommend readers
refer to~\cite{DBLP:conf/dagm/WenzelWYCKSZC20} for more details about the challenged dataset. As is expected, 
Our method outperforms COLMAP by a large margin in terms of both accuracy and efficiency. In the Old Town scene, COLMAP failed to 
reconstruct on sequence recording\_2020-10-08\_11-53-41 and sequence recording\_2021-02-25\_12-34-08 (we use - to denote the failed cases). 
As the two sequences contain severe motion blur and tunnels in images, which makes them very challenging to reconstruct. However, our method 
is also robust to these scenes since it can robustly fuse different sensor data.

\end{document}